\def\year{2020}\relax
\documentclass[letterpaper]{article} % DO NOT CHANGE THIS
\usepackage{aaai20}  % DO NOT CHANGE THIS
\usepackage{times}  % DO NOT CHANGE THIS
\usepackage{helvet} % DO NOT CHANGE THIS
\usepackage{courier}  % DO NOT CHANGE THIS
\usepackage[hyphens]{url}  % DO NOT CHANGE THIS
\usepackage{graphicx} % DO NOT CHANGE THIS
\usepackage{graphicx}  % DO NOT CHANGE THIS
\usepackage{wasysym}
\usepackage{multirow}
\usepackage{comment}
\usepackage{setspace}
\usepackage{graphicx,txfonts}
\usepackage{pgfplots}
\usepackage{tabularx}
\usepackage{booktabs}
\usepackage{todonotes}
\usepackage{amssymb}
\usepackage[inline]{enumitem}
\usepackage{xcolor}
\newcommand{\red}[1]{\textcolor{red}{#1}}

\newcommand{\steven}[1]{\textcolor{blue}{#1}}

\title{Learning Cross-lingual Word Embeddings from Twitter via Distant Supervision}
\author{Jose Camacho-Collados$^{\diamondsuit}{}^*$ ~ Yerai Doval$^{\heartsuit}$\thanks{Authors marked with an asterisk contributed equally.} ~ Eugenio Mart\'{i}nez-C\'{a}mara$^{\spadesuit}$ \\ \Large ~\textbf{Luis Espinosa-Anke$^{\diamondsuit}$ ~ Francesco Barbieri$^{\clubsuit}$ ~ Steven Schockaert$^{\diamondsuit}$}\\ %\Large \textbf{AAAI Style Contributions by
$^\diamondsuit$ School of Computer Science and Informatics, Cardiff University, United Kingdom  $^\clubsuit$ Snap Inc., USA  \\ $^\heartsuit$ Escola Superior de Enxe\~{n}ar\'ia Inform\'atica, Universidade de Vigo, Spain
$^\spadesuit$ DaSCI Institute, University of Granada, Spain 
$\quad$  \\
{  $^{\diamondsuit}$\{camachocolladosj,espinosa-ankel,schockaerts1\}@cardiff.ac.uk}, { $^{\clubsuit}$fbarbieri@snap.com} \\
{  $^{\heartsuit}$yerai.doval@uvigo.es},
{  $^{\spadesuit}$emcamara@decsai.ugr.es},
\\
}
 
\begin{document}

\maketitle

\begin{abstract}
Cross-lingual embeddings represent the meaning of words from different languages in the same vector space. Recent work has shown that it is possible to construct such representations by aligning independently learned monolingual embedding spaces, and that accurate alignments can be obtained even without external bilingual data. In this paper we explore a research direction that has been surprisingly neglected in the literature: leveraging noisy user-generated text to learn cross-lingual embeddings particularly tailored towards social media applications. 
While the noisiness and informal nature of the social media genre poses additional challenges to cross-lingual embedding methods, we find that it also provides key opportunities due to the abundance of code-switching and the existence of a shared vocabulary of emoji and named entities. Our contribution consists of a very simple post-processing step that exploits these phenomena to significantly improve the performance of state-of-the-art alignment methods. 
\end{abstract}

\section{Introduction}

%\begin{persian}
%سی
%\end{persian}

Twitter provides a wealth of uncurated text \cite{derczynski2013twitter} and has been found to constitute
%exhibits %a number of 
%linguistic, communicative and situational properties that make it 
a valuable source for developing natural language processing (NLP) systems %\cite{martinezcamara2014,felbo2017using,reyes2012humor}.
in, for example, sentiment analysis \cite{martinezcamara2014}, sarcasm detection \cite{felbo2017using} or humour and irony modeling \cite{reyes2012humor}. 
%Linguistic specificities of Twitter include %character limit 
%\steven{length} restrictions, the use of pseudowords \steven{such as} \#\textit{hashtags} or @\textit{mentions}, and emoji. 
Given their abundance and multilingual nature, we argue that tweets are a powerful but surprisingly neglected source for learning cross-lingual vector representations of words (henceforth, cross-lingual embeddings). 

%\begin{comment}

Cross-lingual embeddings are the result of mapping two or more monolingual word embedding spaces into a shared vector space %where pairwise word translations 
in which words and their translations are represented by similar vectors. Along with %transparent
obvious applications in, for example, machine translation \cite{artetxe2018iclr,lample2018unsupervised,lample:2018emnlp}, cross-lingual embeddings also constitute a major step forward towards knowledge transfer between languages \cite{ruder2019survey}, usually having English as source or pivot. %Specifically, they have proven to be useful in downstream applications such as information retrieval \cite{vulic2015monolingual}, entity linking \cite{tsai2016cross} and text classification \cite{mogadala2016bilingual}, while at the same time providing improvements in multilingual NLP tasks such as machine translation~\cite{zou2013bilingual}. 
%Typically, these %embeddings are learned via linear transformations 
%mappings are learned as linear transformations with the help of external bilingual data \cite{mikolov2013exploiting,faruqui2014improving,dinu2015improving,xing2015normalized,vulic2016bilingual,levy2017strong},  %dinu2015improving,artetxe-labaka-agirre:2017:Long,doval:meemiemnlp2018
Several recent approaches have shown that accurate mappings %are indeed possible %can actually be found 
can indeed be learned
with minimal amounts of supervision, to the point that external bilingual data may no longer be needed \cite{conneau2018word,artetxe:acl2018,xu2018crosslingual}. % (see Section \ref{secRelatedWork}).
However, previous work has mostly focused on controlled or noise-free environments, reporting results from using clean and comparable corpora as source. % (e.g.\ obtained from different language editions of Wikipedia). %so that no external bilingual data is required \cite{conneau2018word,artetxe:acl2018}.
% In contrast, we mainly contribute with a method that can leverage a minimal signal of supervision from Twitter corpora that performs a competitive mapping between monolingual embedding spaces.
%\end{comment}
In this paper we make a case for the potential (and discuss the limitations) of social media %(and Twitter in particular) 
data for learning %high quality 
cross-lingual embeddings, thus parting ways with the traditional `noise-free' setting explored in most recent literature.

%In this paper 
%\steven{In particular,} we argue that \steven{Twitter offers important opportunities} %leveraging noisy user-generated text 
%has a number of important advantages 
%for learning cross-lingual embeddings.
%(i.e., word embeddings in different languages which share the same vector space). %without relying on external bilingual data. % that can be exploited.
%in cross-lingual settings. 
%First, %and foremost, 
In monolingual settings, it has already been shown that word embeddings trained on Twitter lead to increased performance in social media NLP tasks \cite{tang2014learning,godin2015multimedia,Yang2018}.
%, e.g., PoS tagging or named entity recognition \cite{godin2015multimedia} %, as well as
%\steven{and} modelling sentiment \cite{tang2014learning}. 
One of the main reasons 
%lies in the fact 
is that such embeddings cover a much wider range of slang terms and neologisms, and therefore provide a more faithful snapshot of the particularities of the language used in social media. %sites. 
%Moreover, 
Twitter-specific cross-lingual embeddings can thus also be expected to provide solid grounds for cross-lingual social media NLP tasks. In this paper, we demonstrate that this is indeed the case for, specifically, word translation and cross-lingual sentiment analysis, where we use data for English to train classifiers for other languages. %learning sentiment classifiers for languages other than English 
%, for which training data may be scarce or even non-existent, by 
%by exploiting English-language training data. 

%Another important advantage of Twitter \steven{for learning cross-lingual embeddings} 
%, which is specific to the cross-lingual setting, 
%is the pervasiveness of \textit{code-switching}, i.e.\ the phenomenon where words from more than one language are used within the same tweet. As our experiments demonstrate, this allows for a very effective way of integrating word embeddings from different languages.
%In fact, these Twitter-specific cross-lingual vector spaces are by nature more sensitive to \textit{code-switching}, where words in more than one language are used interchangeably, a pervasive phenomenon in social media.

 %Such cross-lingual embeddings could also play a key role in \emph{code-switching} settings%scenarios such as Twitter
%, where words in more than one language are used interchangeably.
%NLP tools that need to deal with tweets that mix more than one language, a practice known as \emph{code-switching}.

Another crucial advantage of Twitter
%In addition to the above, a more general motivation for our approach 
is that %multilingual Twitter data is peppered with a significant number of shared tokens.
it is peppered with a significant number of tokens\footnote{We use \textit{token} as umbrella term covering anything from a word to an emoji, or any other social media textual artifact.} that are shared across different languages. This is relevant, as previous work has demonstrated that the shared meaning of numerals can be exploited for effectively learning cross-lingual embeddings in a self-supervised fashion \cite{artetxe-labaka-agirre:2017:Long}. For instance, we can assume that the embedding for `5' will embody similar properties in, e.g., English and Spanish. We can also assume that embeddings for emoji obtained from tweets in different languages will generally represent the same or a very similar meaning \cite{barbieri2016cosmopolitan}. %\steven{In this paper, we extend this idea by also considering emoji, since more often than not,} embeddings for emoji obtained from tweets in different languages \steven{should} represent the same or a very similar meaning \cite{barbieri2016cosmopolitan}. 
Finally, and most importantly, we also take advantage of the fact that many words in non-English tweets have an exact counterpart in English, which can be attributed to code-switching and to the presence of \textit{interlingual homographs}\footnote{Interlingual homographs can be defined as words written identically in two or more different languages.}, including many named entities. %, and the fact that named entities are usually not translated. 
%We demonstrate that numerals, emoji and  interlingual homographs\footnote{Words written identically in two or more different languages.} provide a solid cross-lingual lexical pivot which, 

We exploit this vocabulary of shared tokens across tweets from different languages to implement a very simple post-processing technique,
%These observations are put into practice in this paper via a very simple post-processing technique,  
which maps identical tokens from different languages to the same vector in the cross-lingual embedding space. Clearly, it is overly simplistic to assume that two words from different languages have the same meaning simply because they are spelled in the same way, and even emoji sometimes have language-specific meanings.  Surprisingly, however, we find that such a simple %and heuristic 
post-processing strategy nonetheless leads to substantial performance gains
%superposes these identical words to a common point in the space, %consisting in computing the mean embedding for each pair of these words, 
%promising results 
in the tasks of %\steven{the important task of 
word translation and cross-lingual sentiment analysis. 

Pre-trained monolingual and cross-lingual embedding models for all languages explored in this paper (i.e., English, Spanish, Italian, German and Farsi) are available at \url{https://github.com/pedrada88/crossembeddings-twitter}.\footnote{This repository will be updated with embeddings for additional languages (Finnish and Japanese are already available).}
 %}. % without the need of training data in the target language. 
%To further analyze our approach, we also test it in the standard intrinsic task of word translation, where we find it to be particularly effective for closely-related language pairs.
%the greatest improvement in performance is obtained in closely-related language pairs and languages sharing the same alphabet. 

\section{Motivation}

Learning unsupervised cross-lingual embeddings (i.e. without the need for any kind of external supervision) has become one of the most prominent tasks in NLP in recent years \cite{ruder2019survey}, both as a research challenge in itself  %\todo{What does this refer to? I don't think ``theoretical motivation'' is the right term (at least, I'm not aware of any theoretical papers on this topic).}
and for its potential for transferring knowledge across languages. Such cross-lingual word embeddings have already proved effective in cross-lingual NLP tasks, typically after being trained on standard corpora such as Wikipedia \cite{artetxe-labaka-agirre:2017:Long,conneau2018word,glavas-etal-2019-properly}. However, cross-lingual resources specifically tailored to the colloquial nature of social media are scarce, which constitutes the main motivation for this paper. %lies in the need for multilingual resources in social media. 
We argue that unsupervised cross-lingual embeddings are a highly appealing resource in this regard. %. They are a particularly attractive option in social media environment\luis{s}
%They enable the development of multilingual technologies without requiring any form of (manual or automatic-crafted) supervision
They enable the development of multilingual technologies without requiring any form of manual supervision, which usually ranges from parallel data to word translations, both of which are generally lacking in the social media domain and especially in less-resourced languages. Crucially, this technology can have an impact in applications with direct societal implications, e.g., when natural disasters hit areas where people speak a low-resource language \cite{imran2018processing}. Despite the success of unsupervised cross-lingual embeddings, their effectiveness in social media have remained untested so far. In this paper, therefore, we compile Twitter corpora in five different languages (English, Spanish, Italian, German and Farsi) and provide an extensive analysis of their resulting cross-lingual word vector mappings.

%in this setting

We found that the effectiveness of existing state-of-the-art cross-lingual embeddings methods in the social media setting is limited%\todo{Unclear. Are you suggesting that the cross-lingual embeddings we can get from Twitter are not that useful after all?}
, but discovered that using \textit{free} supervision such as identical tokens can be a promising workaround. After analyzing the nature of these identical tokens (see Section \ref{signals} and the ablation test in Section \ref{ablation}), we put forward a simple post-processing step which causes a sharp performance increase for all languages and tasks we considered, both intrinsic and extrinsic, over state-of-the-art unsupervised and weakly-supervised cross-lingual embedding models. The motivation for this post-processing technique (see Section \ref{superposition}) is to exploit some specific features of social media text, namely its code-switching nature and the occurrence of special interlingual tokens such as numerals and, especially, emoji. These features are not usually present in other different types of corpora but, as we will show, they turn out to be powerful bilingual signals in cross-lingual embedding learning.

\section{Related Work}\label{secRelatedWork}

\subsection{Cross-lingual word embeddings}

Cross-lingual embeddings are becoming increasingly popular in NLP \cite{upadhyay2016cross,ruder2019survey}, especially since the recent introduction of models requiring almost no supervision \cite{mikolov2013exploiting,faruqui2014improving,xing2015normalized,smith2017offline,artetxe-labaka-agirre:2017:Long,doval:meemiemnlp2018}. These models have shown to be highly competitive compared to %perform equally or \steven{even} better than 
fully supervised baselines (which are typically trained on parallel corpora). %requiring substantial supervision, which is usually provided in the form of parallel corpora. %large amounts of supervision such as parallel corpora. %, enabling cross-language knowledge transfer without traditional supervision signals such as parallel corpora. 

Despite their effectiveness, these recent models still need 
%suffer from the need of requiring 
some form of supervision signal, which often takes the form of a bilingual dictionary. This limitation motivated 
the emergence of fully unsupervised models, based on,
%a complementary fully unsupervised direction base on, 
among others, adversarial training \cite{zhangcrosslingual:emnlp2017,conneau2018word,xu2018crosslingual,chen2018unsupervised}. %However, experimentally, the experimental results described may be seen as somewhat limited, as they were performed only on close languages and using comparable corpora (i.e., Wikipedia). 
%fact, in an evaluation study on the bilingual dictionary induction task, 
However, as shown by \citeauthor{sogaard2018limitations} \shortcite{sogaard2018limitations}, some of these fully unsupervised methods (e.g., \citeauthor{conneau2018word} \shortcite{conneau2018word}) may 
%suffer from brittleness 
be brittle when dealing with different types of languages and corpora. %, finding that the performance dropped significantly in more challenging settings, such as using different corpora and morphologically-rich languages. 
%However, their evaluation was focused on this particular model, i.e. \citeauthor{conneau2018word} \shortcite{conneau2018word}, corpora of different nature (Wikipedia and Europarl) and on the bilingual dictionary induction task only. 
In a parallel direction, \citeauthor{artetxe:acl2018} \shortcite{artetxe:acl2018} proposed an alternative unsupervised model for learning cross-lingual embeddings, based on a similarity-based dictionary initialization and a linear transformation. While this approach proved to be more robust, and can even surpass supervised models exploiting synthetic or external bilingual dictionaries \cite{mikolov2013exploiting,xing2015normalized,smith2017offline}, they only considered standard corpora.

In this paper we evaluate some of the most prominent cross-lingual embedding models in the more challenging setting of social media. Our evaluation shows that unsupervised models often struggle with noisy user-generated text, and the resulting aligned spaces seem to perform poorly in standard evaluation benchmarks (both intrinsic and extrinsic). 

\subsection{Cross-lingual sentiment analysis}

As with most NLP tasks, the availability of training data and linguistic resources for sentiment analysis (SA) is generally skewed towards English, which motivates the creation of cross-lingual SA systems. %, which may be applied to non-English texts. 
However, most existing work in cross-lingual SA is built upon \begin{enumerate*}[label=(\arabic*)] \item machine translation systems \cite{salameh2015,zhou2016}; \item parallel  \cite{meng2012,chen2018adversarial} or comparable corpora \cite{Rasooli2018}; \item synthetic corpora developed with documents written in the source and the target language \cite{vilares2017}; or \item bilingual lexicons \cite{barnes-etal-2018-bilingual}.
\end{enumerate*} %\citet{Rasooli2018} avoided the constraint of using machine learning systems or parallel corpora by instead using comparable corpora.
Consequently, all these works depend on the availability of annotated data or the quality of off-the-shelf machine translation systems, which are generally ill-suited for social media text. In contrast, the approach we consider in this paper effectively enables zero-shot cross-lingual transfer in sentiment analysis without the need for external bilingual resources.

%\red{New related paper: \cite{chen2018adversarial}}
%martinezCamara2014b,

\section{Learning Cross-lingual Embeddings} % without External Bilingual Signals}
\label{methodology}

Most approaches for learning cross-lingual embeddings without parallel corpora make use of standard pre-trained monolingual vectors.
%and then apply several transformations from them.  %\cite{zhangcrosslingual:emnlp2017,conneau2018word,artetxe:acl2018}. 
%Monolingual word embeddings 
%\steven{that} can be learned with common models \steven{such as Skip-Gram}, %\cite{Mikolovetal:2013} 
%GloVe or FastText. %\cite{pennington2014glove}. 
%In the context of user-generated data, however, FastText \cite{bojanowski2017enriching} emerges as the most natural choice, as this model also exploits subword units and is therefore more robust to misspellings. 
%Once these monolingual embeddings are trained, bilingual signals are
These are mapped onto a shared cross-lingual space, usually with the help of external bilingual dictionaries. %The minimum amount of supervision required for this alignment step is a bilingual dictionary.
%which can be used from external resources or constructed automatically. 
As an alternative, in this paper we consider automatically acquired dictionaries. % that are constructed fully automatically.
%However, as we show in this paper, strong performance can be achieved even in cases where this dictionary is itself constructed in a fully automated way, thus alleviating the need for having external resources available.
%and techniques to construct such dictionaries are presented in the following section.
In Section \ref{signals}, we discuss how these dictionaries can be constructed from Twitter data. These dictionaries will then be used as the supervision signal for well-known state-of-the-art methods, which are briefly recalled in Section \ref{alignments}. Finally, we introduce a simple post-processing step which drastically improves performance in different benchmarks (Section \ref{superposition}). 

%Learning cross-lingual embeddings without external resources usually consists of two main steps, where the first step consists of learning monolingual embeddings. Then, an alignment strategy is used to map the monolingual embeddings to a common bilingual vector space.

\subsection{Automatic creation of a bilingual dictionary}
\label{signals}

%As far as models without exploiting external bilingual data are concerned, 

%there are different implicit signals that can be exploited as distant supervision. 
There are two main approaches to automatic dictionary construction from monolingual corpora: by distant supervision or by relying on the distribution of monolingual embeddings. In our method, we will rely on distant supervision signals from Twitter. However, let us first briefly introduce the latter ``fully unsupervised'' methods.

\smallskip
\noindent\textbf\textbf{\textbf{Unsupervised (distributional).}} %Other methods do not take external dictionaries from identical words as the previous examples but rather create a synthetic one 
Approaches from this class construct a dictionary by exploiting the %similarity 
distribution of monolingual embeddings. There are two prominent methods that rely on this intuition: \citeauthor{artetxe:acl2018} \shortcite{artetxe:acl2018} exploit the structural similarity of %the
monolingual embeddings, specifically, %by relying on the assumption that%
the fact that cross-lingual synonyms have close similarity distributions across different languages. \citeauthor{conneau2018word} \shortcite{conneau2018word}, on the other hand, learn this initial bilingual dictionary through adversarial training.

\smallskip
\noindent\textbf{\textbf{Distant supervision (identical tokens). }} To construct a synthetic bilingual dictionary in an automatic fashion, we rely on the following intuition: whenever a token appears in both monolingual corpora, we assume it has the same meaning. %; our initial dictionary only contains these trivial translations. 
In other words, our dictionary only contains trivial entries, where a word is equal to its (presumed) translation.
%We consider the following types of tokens.
These identical tokens can be split into the following three types:
%We will in particular analyze the following types of tokens: %These identical words can be categorized in three types, which we analyze separately:

 %\begin{itemize}%[noitemsep]
 
 \smallskip
 \noindent\textbf{(i) Numerals:} Given their extensive usage, %even using different alphabets, 
 Arabic numerals constitute a ubiquitous cross-lingual distant supervision signal. %an interesting source to automatically build bilingual dictionaries. 
 They were first leveraged by \citeauthor{artetxe-labaka-agirre:2017:Long} \shortcite{artetxe-labaka-agirre:2017:Long}. %, whose iterative approach proved competitive with respect to %supervised methods making use of 
 %external bilingual dictionaries.
 
 \smallskip
\noindent \textbf{(ii) Emoji:} Emoji are ideograms depicting people, objects and scenes \cite{cappallo2015image2emoji}, which co-exist with words in social media communication% are included in social media communication alongside words to convey messages%
 . While some emoji preserve cultural differences, they have been shown to share similar meaning across languages and countries \cite{barbieri2016cosmopolitan}. One of their potential advantages with respect to numerals, in addition to their prevalence
 %universal usage 
 in social media, is their 
 %variety
diversity, as there are emoji for a wide range of domains such as medicine (\includegraphics[height=0.32cm,width=0.32cm]{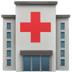}), sports (\includegraphics[height=0.32cm,width=0.32cm]{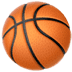}), business (\includegraphics[height=0.32cm,width=0.32cm]{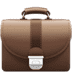}) or geography (\includegraphics[height=0.32cm,width=0.32cm]{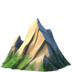}). %, to name %just but a few. 
Emoticons such as smileys, e.g., \texttt{:-)}, provide a similar bilingual signal.
 
 \smallskip
 \noindent\textbf{(iii) Shared words}: 
 %\textbf{Interlingual homographs}: 
 %In addition to the numerals and emoji, there may be other words which share the same token across languages. 
English words are often used by non-English speakers in spontaneous communication in social media.
 %English words are often used in non-English tweets, and vice versa. 
 This phenomenon is particularly common % especially prevalent %
 in languages that are related to or which share their alphabet with English, where vocabularies of shared words may arise due to the existence of interlingual homographs\footnote{Clearly, there are examples of words which are written in the same way in two languages, but which have a different meaning. For instance, the correct English translation of the Spanish word \textit{sensible} is \textit{sensitive}, not \textit{sensible}. %(which has a different meaning in English).  %both a Spanish and an English word but their corresponding meanings differ: \textit{sensible} in Spanish would be translated as \textit{sensitive}.} and references to named entities,
Nonetheless, such a na\"{i}ve assumption %that interlingual homographs have the same meaning 
proves to be indisputably 
 %valid
helpful.} % in this case%leads to an improved performance
 or code-switching environments. Even in more distantly related languages, English words are used in the form of many borrowed and loan words, especially in digital communication. 
 %\stevenNote{Why don't we talk about code-switching anymore here? We need to be consistent and either not talk about it at all (e.g.\ in the introduction) or also mention it here. Added} 
 %, although this also occurs for more distinct languages due to code-switching and the universal usage of English as a vehicular language.
 %In this group of identical words we may find positive instances such as loanwords or entities like companies or people names which may not change their writing from language to language.
 %where they may co-exist on a code-switching context (most languages share this phenomenon with the English language). 

\subsection{Alignment strategies}
\label{alignments}
Various methods have been proposed for aligning two monolingual embedding spaces. 
%In order to align pairs of monolingual embedding spaces, various heterogeneous approaches have been proposed. 
Two recent methods in particular have obtained outstanding results in both unsupervised and semi-supervised settings: MUSE \cite{conneau2018word} and VecMap \cite{artetxe2018generalizing}. %, which we use as our base model. 
Recall that the seed supervision signal required for these methods comes in the form of a bilingual dictionary, which may be external or automatically generated. % (by distant supervision or fully unsupervised) as explained in the previous section. 
These two methods are similar in that they learn an orthogonal linear transformation which maps one monolingual embedding space into the other.
%These may be used in a supervised or unsupervised fashion, and at their core both methods learn an orthogonal linear transformation which maps one of the monolingual embedding spaces into the other.
In VecMap this is done using SVD, while MUSE uses Procrustes analysis. %followed by de-whitening, re-weighting and dimensionality reduction of the resulting embeddings.
VecMap applies this approach in an iterative fashion, where at each step the previously used bilingual dictionary is extended based on the current alignment.
%The supervision signal required for the initial alignment comes in the form of a bilingual dictionary, which can be obtained automatically and may be as small as 25 entries long. 
It is also worth noting that after the initial orthogonal transformation, VecMap fine-tunes the resulting embeddings by giving more weight to highly correlated embedding components, improving its performance in word translation.
%In a recent work, \citeauthor{artetxe:acl2018} \shortcite{artetxe:acl2018} show how the initial bilingual dictionary can be automatically obtained by analyzing the relative distribution of words in their source domains.
%, which nonetheless may be obtained in a fully unsupervised manner as explained in the previous section. %In order to learn this bilingual dictionary, VecMap applies a method that lies on the assumption of cross-lingual synonyms having close similarity distributions in different languages. %VecMap can start with a small seed dictionary and obtain larger dictionaries by applying the learned mappings in an iterative process.
%On the other hand, this seed dictionary may be given by the user or obtained automatically from the training corpora using pairs of words with similar distributions.
%
 %and apply its transformation to obtain a new one. % which we also evaluate in this paper. 
%MUSE, on the other hand, learn this initial bilingual dictionary through adversarial training. \steven{Both} methods can also be used in semi-supervised settings.
%On the other hand, MUSE can also learn its bilingual dictionary in an unsupervised fashion through adversarial training. %MUSE learns a bilingual dictionary through adversarial training, although it can also be used in a semi-supervised setting.

%As a complement to these methods, 
Finally, let us refer to \citeauthor{doval:meemiemnlp2018} \shortcite{doval:meemiemnlp2018}, who recently proposed a method which extends VecMap and MUSE with a post-processing step. This method consists in applying an additional linear transformation, learned by linear regression on the translation pairs from external bilingual dictionaries. In this way, %in the multilingual embedding space 
cross-lingual synonyms are mapped to their corresponding average embedding. Note that this dictionary can again be obtained through distant supervision, although this was not explored in \citeauthor{doval:meemiemnlp2018} \shortcite{doval:meemiemnlp2018} or its extension \cite{doval2019meemi}.
%Note that even though in their initial experiments an external bilingual dictionary was leveraged, this dictionary may be obtained through distant supervision as well.% as explained in the previous section.

%Furthermore, we experiment with a variation of this method where we give different weights to the words in the translation pairs based on their frequencies in their respective corpora.
%In this way, we give more weight to those words with better representations in their corresponding corpora.

%\subsection{Superposition of cross-lingual embeddings in the mean}
\subsection{Averaging cross-lingual embeddings}
\label{superposition}

We put forward a simple post-processing step inspired by \citeauthor{doval:meemiemnlp2018} \shortcite{doval:meemiemnlp2018}. However, in contrast to the latter method, which modifies the vector representations of all words, we simply replace the representations of the words in our synthetic dictionary by the average of their initial vector and the initial vector of their presumed translation, leaving all other vectors unchanged. 
%which we may summarize as follows. Instead of applying a single linear transformation to all the word vectors in the cross-lingual embedding space, we simply average those that are present in a reference bilingual dictionary, leaving the remaining vectors unchanged. 
In our experimental results, we show that, surprisingly, this simple approach leads to substantially better results than those obtained by competing baselines. 
%to these alignment strategies. Instead of applying a linear transformation to the aligned cross-lingual embeddings as in \citeauthor{doval:meemiemnlp2018} \shortcite{doval:meemiemnlp2018}, our proposal is to simply move the word embeddings present in a given bilingual dictionary to their average. %raw mean embeddings as the new cross-lingual embeddings for those words in a bilingual dictionary.
%Given that the rest of the words not present in the dictionary would remain unchanged, it is essential to learn a base cross-lingual model (e.g. VecMap) prior to the application of our method, as well as having with a reasonably well-sized bilingual dictionary, which can be achieved through distant supervision.
Our method crucially relies on the availability of a sufficiently large bilingual dictionary. In this regard, one of the main contributions of this paper is showing that suitable dictionaries can be obtained automatically from Twitter corpora. %, which in turn makes it possible to fully exploit this \steven{distantly supervised} approach. % using distant supervision.

In addition to this vanilla averaging method, we also consider a variant in which the average is weighted by %\steven{relative} 
frequency: 
\begin{equation}
\vec{\mu}_{w_1,w_2}=\frac{f_1 \vec{v}_1+f_2 \vec{v}_2}{f_1+f_2}
\end{equation}

\noindent where $f_1$ and $f_2$ are the number of occurrences of the tokens $w_1$ and $w_2$ in their corresponding monolingual corpora,
%, which may have advantages or disadvantages. With the absolute frequencies the model would rely on the word which should have the better embedding as its frequency is higher, while with the relative frequency the model would consider both corpora as equally relevant, irrespective of their size.}
and $\vec{v}_1$ and  $\vec{v}_2$ represent the embeddings of $w_1$ and $w_2$ in the cross-lingual vector space.\footnote{$f_1$ and $f_2$ may be either absolute or relative frequencies. In our case we did not find noticeable differences given that all monolingual corpora were of comparable size. %\todo{Does this mean ``given that all monolingual corpora were of comparable size''?} 
} % simply used absolute frequencies as there were no notable performance differences given the relatively similar corpora size.} 
%the movement to the average will be more pronounced when the word is not sufficiently represented in a given language (and therefore its embedding is expected to be of lower quality), 
The main intuition behind this alternative is that even when a word occurs in tweets from both languages, it may still be underrepresented in one of them. This would be the case, for instance, if in one of the languages the word were only used in a code-switching context, or simply because of it being less prominent due to cultural or geographical differences.
%which could be a strong indication of its usage as a loan word, in a code-switching context, or simply less used due to cultural or geographical differences. 
For instance, the word \textit{NFL}, which stands for \textit{National Football League} in the United States is also used in Spain, but much less frequently. We can thus expect that its Spanish embedding will be less accurate than the English one. Therefore, in this case it would make sense to give more prominence to the English vector. %, which we can accomplish by using the weighted average. 
We will use \textit{Plain} and \textit{Weighted} to refer to our standard and weighted averaging strategies respectively.

%Our unweighted averaging postprocessing method will be referred to as \textit{Plain} henceforth, and the weighted one as \textit{Weighted}.

%move its Spanish embedding towards the English one, which would further approach the spaces while intuitively improving the Spanish representation for that word.

%TODO: using cognates for this may be possible thanks to the code-switching...

\section{Evaluation}
\label{evaluation}
%In this section we explain the common experimental setting to train cross-lingual embeddings for all models.

%\subsubsection{Languages}
%Our main focus \steven{is on analyzing} 
%on performing an analysis of 

We analyze the performance of cross-lingual word embeddings in the context of Twitter corpora, focusing in particular on the effectiveness of our post-processing method.
%multilingual settings. To this end, we perform extensive evaluation in various benchmarks, 
%\steven{We consider} both intrinsic and extrinsic \steven{benchmarks}. %specifically on cross-lingual sentiment analysis . %We also perform an ablation study is provided in Section \ref{ablation}.
%, however, let us 
First, however, let us describe %provide further details on 
the setting for cross-lingual embedding training. 

%sentiment analysis. This is a natural task for our setting, since sentiment analysis requires embeddings trained on Twitter data (or closely related corpora), due to the importance of capturing informal language and specific phenomena from social media. %such as emoji. 
%After intrinsically evaluating our method in the word translation task in Section \ref{analysis}, we present further results on the cross-lingual sentiment analysis in Section \ref{sentanalysis}.
 %Finally, an ablation study is provided in Section \ref{ablation}.
 %our experimental set-up.
%We test the cross-lingual embeddings in the cross-lingual sentiment analysis task, using English as source language and Spanish and Italian as target languages. 
%two romance languages which are relatively close to the English language (Spanish and Italian) and share the same alphabet. %For the analysis (Section \ref{analysis}) we additionally include two far agglutinative languages (Japanese and Farsi) which use a different alphabet. 

%\subsection{Cross-lingual embeddings training}
%\label{experimentalsetting}

%In this section we explain the common experimental setting to train cross-lingual embeddings for all models.

\medskip
\noindent\textbf{Corpus compilation}
We collected five monolingual Twitter corpora between October 2015 and July 2018.\footnote{While all tweets were downloaded between these two months for all corpora, the Farsi set contains tweets from a shorter period.} 
These corpora were independently gathered using geolocalized tweets which were tagged with specific languages:\footnote{We relied on Twitter language identification procedure for gathering the language-specific tweets.} United States (English), Spain (Spanish), Italy (Italian), Germany (German) %Japan (Japanese) 
and Iran (Farsi). To encourage more tweet diversity, only a maximum of twenty tweets per user were retained. After preprocessing (tokenization and duplicate removal), the final corpora consisted of 21,461,242 tweets for English, 10,122,550 for Spanish, 4,546,508 for Italian, 7,905,827 for German %19,501,589 for Japanese 
and 3,724,602 for Farsi. All languages contained more than 60M tokens overall and more than 1M unique tokens, with English being the largest among the five languages considered.
%The resulting monolingual corpora ranged from around %2.5 
%3M tweets for Farsi, which is the shortest corpus, to 20M tweets for English, the largest in number of tweets. 
Table \ref{tab:corpora} summarizes the main statistics (overall number of tweets and tokens, and number of unique tokens) of all Twitter language-specific corpora used in our experiments.

%\begin{comment}

\begin{table}
\begin{center}
%\renewcommand{\arraystretch}{1.15}

%\scalebox{0.7}
%\small
%{
%\setlength{\tabcolsep}{4.0pt}
%\scalebox{0.91}{ 
%\resizebox{\columnwidth}{!}{
\begin{tabular}{@{}lrrr@{}}
%\cline{2-5}
\toprule
{\bf Language} 	& {\bf \# Tweets} 	&	{\bf \# Tokens} 	&	{\bf \# Unique}	\\ 

%\cmidrule(l{4pt}r{4pt}){2-4} \cmidrule(l{4pt}r{4pt}){5-7} 
%\multicolumn{1}{c|}{} &   {\bf F-Measure}  & {\bf F-Measure}   \\ 
%\cmidrule(l{4pt}r{4pt}){1-1} \cmidrule(l{4pt}r{4pt}){2-4} \cmidrule(l{4pt}r{4pt}){5-7} 

%\hline
%SW2V(Wiki) & - &  -	 \\
\midrule
English 	&   21,461,241	&  294,276,603 &  5,499,846 \\
%\hline
Spanish   	&   10,122,550	& 144,394,815 & 3,312,603    \\
%\hline
Italian 	&   4,546,508	& 63,076,614 & 1,601,218   \\   
%\hline
%Japanese 	&   19,501,589	& 423,233,103    \\   
%\hline
German 	&   7,905,827	& 114,545,634 &  2,301,059   \\   
%\hline
Farsi 	&   3,724,602	& 90,288,567 &  1,038,666 \\   
%Farsi 	&   2,574,546	& 62,584,915    \\   

\bottomrule
\end{tabular}
%}
%}
\end{center}
\caption{\label{tab:corpora} Number of tweets and tokens (overall and unique) per corpus.
}
\end{table}
%\end{comment}

\smallskip
\noindent\textbf{Monolingual embeddings}
All comparison systems use the same monolingual embeddings as input. These embeddings were trained on the Twitter corpora described above using FastText \cite{bojanowski2017enriching}. %As mentioned in Section \ref{methodology} 
%which % as word embedding model %instead of Word2vec \cite{Mikolovetal:2013} or GloVe \cite{pennington2014glove} 
FastText was chosen due to its handling of subword units, making it more robust to misspellings as compared to alternatives like Word2Vec \cite{Mikolovetal:2013} or GloVe \cite{pennington2014glove}. %We used the default FastText hyperparameters and a dimensionality of 100.
The monolingual embeddings were trained with FastText's default hyperparameters, fixing the dimension size to 100.

%\subsubsection{Synthetic dictionaries}

\newcolumntype{L}{>{\raggedright\arraybackslash}X}
\begin{table*}[!tb]
\begin{center}
%\scalebox{0.7}
\scriptsize
%\footnotesize
%\renewcommand{\arraystretch}{1.15}
%\resizebox{\textwidth}{!}{

%{
%\setlength{\tabcolsep}{4.0pt}

%\scalebox{0.9}{ 
%\setlength\tabcolsep{1pt}
\begin{tabularx}{\textwidth}
%{@{}l@{\hphantom{0}}l@{\hphantom{0}}r@{\hphantom{0}}r@{\hphantom{0}}r@{\hphantom{0}}r@{\hphantom{0}}r@{\hphantom{0}}r@{\hphantom{1}}r@{\hphantom{0}}r@{\hphantom{0}}r@{\hphantom{0}}r@{\hphantom{0}}r@{\hphantom{0}}r@{\hphantom{1}}r@{\hphantom{0}}r@{\hphantom{0}}r@{\hphantom{0}}r@{\hphantom{0}}r@{\hphantom{0}}r@{\hphantom{1}}r@{\hphantom{0}}r@{\hphantom{0}}r@{}}
{@{}LLr@{\hphantom{0}}r@{\hphantom{0}}rr@{\hphantom{0}}r@{\hphantom{0}}rr@{\hphantom{0}}r@{\hphantom{0}}rr@{\hphantom{0}}r@{\hphantom{0}}rr@{\hphantom{0}}r@{\hphantom{0}}rr@{\hphantom{0}}r@{\hphantom{0}}rr@{\hphantom{0}}r@{\hphantom{0}}r@{}}

\toprule
\multirow{3}{*}{\textbf{Supervision}} 	& \multirow{3}{*}{\bf Model} 	&   \multicolumn{6}{c}{\textbf{EN-ES}}  &   \multicolumn{6}{c}{\textbf{EN-IT}}  &   \multicolumn{6}{c}{\textbf{EN-DE}}  &   \multicolumn{3}{c}{\textbf{EN-FA}}\\
\cmidrule(lr){3-8}
\cmidrule(lr){9-14}
\cmidrule(lr){15-20}
\cmidrule{21-23}
&   &\multicolumn{3}{c}{\textbf{Europarl}} &  \multicolumn{3}{c}{\textbf{Facebook}} &  \multicolumn{3}{c}{\textbf{Europarl}} &  \multicolumn{3}{c}{\textbf{Facebook}}   &   \multicolumn{3}{c}{\textbf{Europarl}} & \multicolumn{3}{c}{\textbf{Facebook}} & \multicolumn{3}{c}{\textbf{Facebook}}\\ 
\cmidrule(lr){3-5}
\cmidrule(lr){6-8}
\cmidrule(lr){9-11}
\cmidrule(lr){12-14}
\cmidrule(lr){15-17}
\cmidrule(lr){18-20}
\cmidrule{21-23}
%\cmidrule(l){21-24}
%& & \multicolumn{3}{c}{\textbf{Europarl (EN-ES)}} &  \multicolumn{3}{c}{\textbf{Facebook (EN-ES)}} &  \multicolumn{3}{c}{\textbf{Europarl (EN-IT)}} &  \multicolumn{3}{c}{\textbf{Facebook (EN-IT)}} & \multicolumn{3}{c}{\textbf{Europarl}} & \multicolumn{3}{c}{\textbf{Facebook}} & \multicolumn{3}{c}{\textbf{Facebook}} \\
    & & P1 & P5 &   P10  & P1 & P5 &   P10  & P1 & P5 &   P10  & P1 & P5 &   P10  & P1 & P5 &   P10  & P1 & P5 &   P10  & P1 & P5 &   P10\\
%\cmidrule(l{4pt}r{4pt}){2-4} \cmidrule(l{4pt}r{4pt}){5-7} 
%\multicolumn{1}{c|}{} &   {\bf F-Measure}  & {\bf F-Measure}   \\ 
%\cmidrule(l{4pt}r{4pt}){1-1} \cmidrule(l{4pt}r{4pt}){2-4} \cmidrule(l{4pt}r{4pt}){5-7} 

%\hline
%SW2V(Wiki) & - &  -	 \\
\cmidrule(r){1-5}
\cmidrule(lr){6-8}
\cmidrule(lr){9-11}
\cmidrule(lr){12-14}
\cmidrule(lr){15-17}
\cmidrule(lr){18-20}
\cmidrule{21-23}
\multirow{2}{*}{Unsupervised}%Un\-supervised}   
& MUSE	&  8.3	& 14.1 & 17.8 &  6.8	& 15.3 & 19.0 &  8.7 &  13.7 &  17.1	& 6.7 & 14.4 &  18.0    &   0.0 & 0.0 &  0.0 & 0.0 & 0.1 &  0.1 &  0.0  &  0.0 &  0.0\\
  	&   VecMap	&  9.5	& 17.0 & 19.4 &  8.1	& 16.4 & 20.4 &  9.2 &  16.9 &  20.9	& 8.8 & 17.0 &  22.3  &    0.2 & 0.4 &  0.9 & 0.1 & 0.4 &  0.5 &  0.0  &  0.0 &  0.0\\
%\midrule
\cmidrule(r){1-5}
\cmidrule(lr){6-8}
\cmidrule(lr){9-11}
\cmidrule(lr){12-14}
\cmidrule(lr){15-17}
\cmidrule(lr){18-20}
\cmidrule{21-23}
%Numerals &   \multirow{2}{*}{Artetxe et al (2017)}	& 47.9	& 51.3 & - 	\\
%Emoji  &  	& 47.3	& 51.7 & -     \\
%\hline
\multirow{5}{*}{Distant}    & MUSE & 2.7	& 5.4 & 7.2 &  2.6	& 5.3 & 7.0 &  3.6 &  9.1 &  12.4	& 4.0 & 10.0 &  13.6    &   1.3 & 2.7 &  3.5 & 1.4 & 3.0 &  4.1 &  0.1  &  0.4 &  0.9\\
   &  VecMap &  10.1	& 17.8 & 21.2 &  8.5	& 16.9 & 21.6 &  9.6 &  17.0 &  21.0	& 9.1 & 16.8 &  21.8 &    3.4 & 6.9 &  9.7 & 2.6 & 6.7 &  9.6 &  0.2  &  0.5 &  1.1   \\
   &   Meemi &  3.7	& 9.7 & 12.4 &  3.9	& 9.1 & 12.0 &  7.9 &  16.2 &  19.1	& 6.7 & 14.0 &  18.2    &   1.9 & 4.5 &  6.8 & 2.0 & 4.1 & 5.9 & 0.1  &  0.3 &  0.4\\
   &   \textit{Plain}	&  \bf 21.1	& 25.9 & 29.6 &  \bf 16.7	& 20.2 & 23.2 &  \bf 20.3 &  28.3 &  33.1	& \bf 22.4 & \bf 31.3 &  \bf 35.7  &   24.0 & \bf 27.3 &  \bf 29.4 & \bf 16.2 & \bf 19.4 &  \bf 21.3 &  \bf 1.3  &  \bf 1.7 &  \bf 2.0\\
   &   \textit{Weighted}	&  20.8	& \bf 28.6 & \bf 33.7 &  \bf 16.7	& \bf 22.8 & \bf 28.5 &  19.4 &  \bf 28.8 &  \bf 33.7	& 21.2 & 28.9 &  33.5    &  \bf 24.3 & 26.2 &  28.6 & \bf 16.2 & 18.0 &  19.9 &   1.2  &  1.5 &  1.8     \\
   
   \cmidrule(r){1-5}
\cmidrule(lr){6-8}
\cmidrule(lr){9-11}
\cmidrule(lr){12-14}
\cmidrule(lr){15-17}
\cmidrule(lr){18-20}
\cmidrule{21-23}
  \multirow{2}{*}{100 pairs}   &   VecMap	&  10.0	& 17.0 & 20.5 &  7.9	& 15.9 & 20.2 &  9.2 &  17.1 &  20.9	& 8.8 & 17.6 &  22.3    &  3.2 & 7.6 &  9.7 & 0.0 & 0.0 &  0.1 &   0.1  &  0.3 &  0.4     \\
   &   MUSE	&  0.6	& 1.7 & 2.5 &  0.4	& 1.1 & 1.9 &  0.6 &  1.6 &  2.3	& 0.3 & 1.1 &  1.8    &  0.3 & 0.5 &  0.5 & 0.1 & 0.3 &  0.6 &   0.1  &  0.3 &  0.4     \\
   
\bottomrule
%\end{tabular*}
\end{tabularx}
%}
%}
%}
\end{center}
\caption{\label{tab:wordtrans} Word translation results on 4 target languages: Spanish (ES), Italian (IT), German (DE) and Farsi (FA). The \textit{Plain} and \textit{Weighted} models are built with our proposed post-processing technique over the base VecMap model.}
\end{table*}

\smallskip
\noindent\textbf{Distant supervision}
As explained in Section \ref{signals}, we automatically extracted bilingual dictionaries of identical tokens %(including numerals, emoji and shared words) 
to be used as supervision for the cross-lingual models. %are automatically extracted via distant supervision. %to be used as supervision for the cross-lingual models. 
This resulted in dictionaries of 122,469 word pairs for English-Spanish, 66,037 for English-Italian, 93,695 for English-German and %85,142 for English-Japanese and %3,290 
6,142 for English-Farsi.

\smallskip
\noindent\textbf{Comparison systems}
We used VecMap \cite{artetxe:acl2018} and MUSE \cite{conneau2018word} to obtain the initial cross-lingual word embeddings, experimenting with their (semi-)supervised and unsupervised settings. In the former case, the supervision came from our synthetic dictionaries of identical tokens. %we used our synthetic dictionaries as supervision.
%As comparison systems we include the unsupervised configurations of VecMap \cite{artetxe:acl2018} and MUSE \cite{conneau2018word}. In addition, we also experiment with these systems in their semi-supervised form, using the synthetic dictionaries of identical tokens as supervision. 
The semi-supervised version of VecMap \cite{artetxe2018generalizing} is used as our base model on which we evaluate two post-processing techniques: Meemi \cite{doval:meemiemnlp2018} and our proposed averaging strategy. Finally, for the sake of completeness, as baseline we also include a version of VecMap which uses external bilingual data in the form of a bilingual dictionary as external supervision. Following \citeauthor{artetxe-labaka-agirre:2017:Long} \shortcite{artetxe-labaka-agirre:2017:Long}, who showed how their semi-supervised model could work with as few as 25 word pairs as supervision, we perform an experiment with 100 word pairs from an external dictionary as supervision.\footnote{These 100 pairs sampled from the Europarl training dictionaries provided by \citeauthor{dinu2015improving} \shortcite{dinu2015improving} and \citeauthor{artetxe-labaka-agirre:2017:Long} \shortcite{artetxe-labaka-agirre:2017:Long}.} For all the baseline systems we followed their official implementations on GitHub.\footnote{VecMap: \url{https://github.com/artetxem/vecmap}; \\ MUSE: \url{https://github.com/facebookresearch/MUSE}; \\ Meemi: \url{https://github.com/yeraidm/meemi}} %For our method we experimented with both plain and weighted averages (see Section \ref{superposition}).

%Weighted %\footnote{Absolute frequencies are used in our experiments with the weighted average. Given the relatively similar size of the corpora of our experiments, the usage of relative or absolute frequencies does not make a noticeable difference. }

\subsection{Intrinsic evaluation: Word translation}
\label{analysis}

\begin{comment}

In this section we further analyze the performance of the cross-lingual embeddings by testing our models in two standard intrinsic tasks, i.e., word translation and cross-lingual word similarity. For these tasks, we can complement our experiments for Spanish and Italian with evaluations for Japanese and Farsi, as representatives of less closely related languages. % for which test sets in these tasks are available. 
Note that we could not include these latter two languages for cross-lingual SA due to a lack of test sets.\footnote{\citeauthor{vaziripour2016analyzing} \shortcite{vaziripour2016analyzing} analyzed the sentiment of political tweets in Farsi but the dataset is not available.}
%The extraction of tweets to compile the corresponding corpora and the training of the cross-lingual embeddings for these two languages followed the same procedure as described in Section \ref{experimentalsetting}.
\end{comment}

%\smallskip
%\noindent\textbf{Word translation}
%\label{bilingualdict}
The task of word translation, or bilingual dictionary induction, %, which is the main task used to evaluate cross-lingual embeddings in the literature, 
%consists of retrieving the correct translation in the target language of a given word in the source language.
consists in retrieving the correct translation in a target language given a word in a source language. %In the context of social media, this can be viewed as a downstream application in itself, as the cross-lingual embeddings can extract translations for concepts not covered in a standard dictionary.

\paragraph{Experimental setting} 
To predict the translation of a word, we return its nearest neighbor from the other language in the cross-lingual embedding space, using cosine similarity.
%The retrieval of a translation given a word is performed through a nearest-neighbor search in the corresponding cross-lingual embedding space using cosine distance.
%As this results in a ranked list of candidates for each word from the source language, . 
The performance is evaluated with the precision at $k$ metric (P-$k$, where $k \in \{1, 5, 10\}$), which is defined in this context %\footnote{In the context of information retrieval, the term precision at $k$ is sometimes used with a different meaning.} 
as the percentage of test instances %pairs (\texttt{word}, \texttt{translation}) 
for which the correct answer is among the $k$ highest ranked candidates.
For this task, we used the standard test sets released by \citeauthor{conneau2018word} and those extracted from Europarl \cite{dinu2015improving,artetxe-labaka-agirre:2017:Long}.

\paragraph{Results}
%Extract nice examples of translations from slang words and jargon (to show the power of social media for this).
As can be observed in Table~\ref{tab:wordtrans}, the plain and weighted averaging methods
%, %applied over the cross-lingual embeddings obtained through VecMap 
%using the dictionary of identical tokens, 
yield the best overall results in Spanish, Italian and German. A similar pattern can be observed for Farsi, although in this case the results are poor overall. The base VecMap embeddings perform better overall in the distantly supervised setting than in the unsupervised setting and even the weakly-supervised setting of 100 pairs. This lends support to the usefulness of synthetically constructed dictionaries with identical tokens in the social media context. The trend contrasts with previous analyses in more standard corpora \cite{vulic2016role}, where this seeding was proved inferior to other strategies. However, this behaviour is not consistent in the case of MUSE, which differs from what was found by \citeauthor{sogaard2018limitations} \shortcite{sogaard2018limitations} on more standard corpora. %\stevenNote{This is misleading as it suggests that \newcite{sogaard2018limitations} already considered synthetic dictionaries with identical tokens. They do indeed :)}.  % suggests  also take advantage of the dictionary of identical words in the \steven{distantly supervised} setting (best results in English-Japanese), as its unsupervised mode generally shows a weaker performance. %The unsupervised model only proves competitive for Japanese, but as we will see in Section \ref{ablation}, even in this case a simple synthetic dictionary based on emojis proves superior. 
%Nonetheless, 
With the exception of English-Farsi, where it fails to generalize\footnote{In this case, VecMap gets stuck in poor local optima, probably due to the non-optimal initialization in this language pair, an issue that was discussed in \citeauthor{artetxe:acl2018} \shortcite{artetxe:acl2018}.}, VecMap outperforms MUSE in both supervised and unsupervised settings. % method of MUSE based on adversarial training and the same model using the dictionary of identical words as supervision.
%For MUSE, the unsupervised setting is its best resource.
% Overall, the unsupervised settings of both tools end up being inferior to %which confirms one of our initial hypothesis? ... NOPE!

Table~\ref{tab:wordtrans} also shows that going from English to %Japanese or, especially, 
Farsi is challenging for all the tested models. This may be attributed to the structural differences between this language and English, and a reflection of cultural differences, which in turn causes a lower prevalence of English words.
%code-switching. 
Indeed, the bilingual dictionary of identical tokens for Farsi is notably the smallest one: 6,142 word pairs against 66,037 for the second-smallest dictionary. %, negatively affecting methods that use no external bilingual knowledge such as the ones studied here.

%Finally, %in spite of the strong results achieved by our proposed post-processing techniques, 
Finally, it is important to highlight 
%some particularities of the test data. Specifically
that the test dictionaries contain a large number of words whose translation is identical to the word itself. Therefore, the fact that the tested methods use synthetic dictionaries which are based on identical tokens might be regarded as giving them an unfair advantage in this task.
%systems exploiting them as supervision may obtain an advantage for free. %, it is important to highlight Table~\ref{tab:dictidentical}.
In particular, this means that the scores obtained for $P@1$ may be artificially high, especially for Spanish and Italian, where the number of words with identical translations is considerable.\footnote{The percentage of identical word pairs in the Facebook test sets are 16.5\% for Spanish, 21.1\% for Italian, 16.0\% for German,  %4.8\% for Japanese, 
and 4.3\% for Farsi. Accordingly, a significant percentage of test pairs are included in our synthetic dictionary: 16.1\% for Spanish, 19.0\% for Italian, 15.7\% for German, %4.1\% for Japanese, 
and 0.4\% for Farsi.} %Therefore, in order to have a broader picture we complement the intrinsic analysis of our cross-lingual embeddings with another intrinsic task (i.e. word similarity). 
However, we should stress that the training dictionaries are obtained automatically from the training corpora, and they were used by all comparison systems in the distantly supervised mode. Furthermore note that they share no connection with the test corpora other than being in the same language.\footnote{In the Europarl test sets word translations were obtained from alignments of the European Parliament proceedings, and therefore reflect a realistic distribution of the languages in that domain.} In what follows we discuss the capability of our post-processing technique by means of a qualitative analysis. %and comparison in a social-media specific extrinsic task, namely cross-lingual sentiment analysis. 
%than the languages considered. %Moreover, at least in the case of the Europarl datasets, the word pairs were obtained from alignments of the European Parliament proceedings, and therefore may reflect a more or less real distribution of the languages.\footnote{To be more precise, these datasets would reflect the distribution in the domain of political proceedings, which is in fact very different from the distributions observed in social media.} %This is reflected in the lower scores of all systems in comparison with the same models applied on more standard corpora like Wikipedia. } 
%All in all, this shows that, while dictionary induction is the most common task to evaluate cross-lingual embeddings, it may not be the most reliable way in similar scenarios to the one presented here, where  %these dataset artifacts can be directly learned by automatic methods. 
%there is a considerable overlap between train and test datasets.
%It is worth noting that this issue is not unique to our present work, having already appeared in (ref) where, in fact, it was not noted or addressed at all.
%Hence, in other to counteract this, 
%\end{comment}

\paragraph{Analysis}

%In Table ?? we can see some interesting example translations. %for slang words and jargon.
We performed an error analysis on our model, examining wrong translations, and found %obtained by the test script, 
that in many cases, the mistranslated word was very similar to the correct translation. %, as it was already noted by \citeauthor{artetxe-labaka-agirre:2017:Long} \shortcite{artetxe-labaka-agirre:2017:Long}.
For example, in our \textit{weighted} model the English verb \emph{requested} is mapped to the Spanish verb \emph{mandado} (\emph{ordered}), and is also near its gold translation, \emph{pedido}. %{\color{red} XXXXX Removed Japanese example, add a similar example for another language? }
%Furthermore, \emph{elephant} is closest to the Japanese word \emph{パンダ} (\emph{panda}) but also close to \emph{ゾウ} \emph{elephant}.
%This means that, while the quantitative results might seem low, a qualitative analysis reveals a generally good behaviour when performing the translation of words from one language to the other, even in far languages like English and Japanese. %with the sole exception of English-Farsi.
%A particular instance of this phenomenon emerges when analyzing the output of the post-processing techniques Meemi and our proposed averaging methods.
%Specially in the former case, %
As far as the baseline post-processing technique is concerned (i.e., Meemi), we can observe substantial drops in the quantitative scores with respect to the base model VecMap. A quick review of the output reveals a general trend of translating source words to target words in the same language (i.e., an English word in the source domain is often translated to some English word which also exists in the target domain). This phenomenon can also be observed for our model. %, although to a lesser extent. 
For example, the five nearest neighbors of the English word \emph{recognize} are also English words from the induced Spanish dictionary: \emph{recognize}, \emph{recognizes}, \emph{acknowledge}, \emph{acknowledged} and \emph{acknowledgement}.
While this is not the intended result for the bilingual dictionary induction task, this reveals a seamless integration of both languages which may partially explain the success of these embeddings in cross-lingual sentiment analysis (Section \ref{sentanalysis}).
%, together with the cross-lingual anchor points developed in our post-processing technique %(unlike Meemi), 
%may partially explain the success of these embeddings in cross-lingual sentiment analysis (Section \ref{sentanalysis}). 

Finally, we performed a more qualitative analysis on the types of translations that cannot be found in standard dictionaries, for which cross-lingual embeddings trained on Twitter are particularly well-suited. Table \ref{tab:slang} shows some examples for translations of selected English slang words and neologisms found in our \textit{weighted} model (top three nearest neighbours according to cosine similarity). From the examples presented, we may draw special attention to \textit{chillax}, a neologism composed of the verbs \textit{chill} and \textit{relax}, which translates also to colloquial ways of referring to the same idea across languages (\textit{relajadito} in Spanish), and perhaps evoking more the notion of coziness in German (\textit{gem\"{u}tlich}). Let us also highlight acronyms like \textit{wth} and \textit{omfg}, whose translations denote surprise, but in an informal register (going as far as translating into swearwords in Farsi, for example). In line with the quantitative results, for Farsi we find mostly noise, but also interesting translations like \includegraphics[width=0.54cm,height=0.35cm]{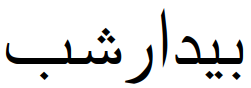} (\textit{sleep}), \includegraphics[width=0.54cm,height=0.35cm]{farsi/chillax-3} (\textit{wake up}) and  \includegraphics[width=0.54cm,height=0.35cm]{farsi/chillax-3} (\textit{night awake}) for `chillax'.

\begin{table}[!h]
\renewcommand{\arraystretch}{1.55}
\setlength{\tabcolsep}{1.25pt}
\resizebox{\columnwidth}{!}{
\begin{tabular}{@{}cccc|cccc@{}}
%\begin{tabular}{@{}c|c|c|c|c|c|c|c@{}}
\toprule
\multicolumn{4}{c|}{\Large \textbf{wth}} & \multicolumn{4}{c}{\Large \textbf{supernerd}} \\ \hline
\textbf{\texttt{ES}} & \textbf{\texttt{IT}} & \textbf{\texttt{DE}} & \multicolumn{1}{c|}{\textbf{\texttt{FA}}} & \textbf{\texttt{ES}} & \textbf{\texttt{IT}} & \textbf{\texttt{DE}} & \multicolumn{1}{c}{\textbf{\texttt{FA}}} \\ \hline
{\Large{pufff}} & {\Large{schifo}} & {\Large{h\"{a}}}&     \begin{minipage}{.1\textwidth}
      \includegraphics[width=0.85\columnwidth,height=0.40cm]{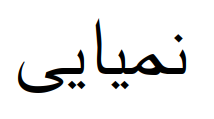}
    \end{minipage} & {\Large{frikifan}} & {\Large{ratman}} & {\Large{lovecrafts}} & \begin{minipage}{.1\textwidth}
      \includegraphics[width=0.85\columnwidth,height=0.40cm]{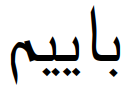}
    \end{minipage} \\ %\hline
{\Large{aggg}} & {\Large{chissene}} & {\Large{h\"{a}\"{a}h}} & \begin{minipage}{.1\textwidth}
      \includegraphics[width=0.85\columnwidth,height=0.40cm]{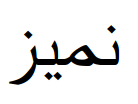}
    \end{minipage} & {\Large{friky}} & {\Large{cinecomic}} & {\Large{trilogie}} & \begin{minipage}{.1\textwidth}
      \includegraphics[width=0.85\columnwidth,height=0.40cm]{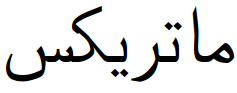}
    \end{minipage} \\ %\hline
{\Large{madremia}} & {\Large{chifo}} & {\Large{n\"{a}\"{a}\"{a}}} & \begin{minipage}{.1\textwidth}
      \includegraphics[width=0.85\columnwidth,height=0.40cm]{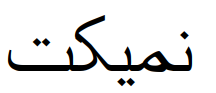}
    \end{minipage} & {\Large{frikie}} & %dell'horror 
    {\Large{fumetti}} & {\Large{gamestar}} & \begin{minipage}{.1\textwidth}
      \includegraphics[width=0.85\columnwidth,height=0.40cm]{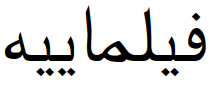}
    \end{minipage} \\ \hline
\multicolumn{4}{c|}{\Large \textbf{chillax}} & \multicolumn{4}{c}{\Large \textbf{omfg}} \\ %\hline
\hline
\textbf{\texttt{ES}} & \textbf{\texttt{IT}} & \textbf{\texttt{DE}} & \multicolumn{1}{c|}{\textbf{\texttt{FA}}} & \textbf{\texttt{ES}} & \textbf{\texttt{IT}} & \textbf{\texttt{DE}} & \multicolumn{1}{c}{\textbf{\texttt{FA}}} \\ \hline
{\Large{relajadito}} & {\Large{rilassando}} & {\Large{entspan}} &
\hspace{0.2cm}\begin{minipage}{.1\textwidth}
      \includegraphics[width=0.60\columnwidth,height=0.40cm]{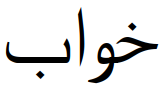}
    \end{minipage} & {\Large{diooo}} & %ommioddio 
    {\Large{mioddio}}
    & {\Large{maaah}} & \begin{minipage}{.1\textwidth}
      \includegraphics[width=0.60\columnwidth,height=0.40cm]{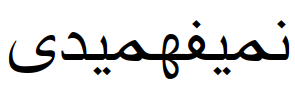}
    \end{minipage} \\ %\hline
{\Large{relaxx}} & {\Large{rilasso}} & {\Large{gem\"{u}tlich}} & \hspace{0.2cm}\begin{minipage}{.1\textwidth}
      \includegraphics[width=0.60\columnwidth,height=0.40cm]{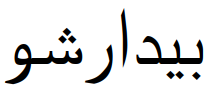}
    \end{minipage} & {\Large{wtfff}} & {\Large{ommiodio}} & {\Large{njaahah}} & \begin{minipage}{.1\textwidth}
      \includegraphics[width=0.60\columnwidth,height=0.40cm]{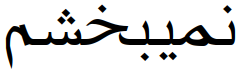}
    \end{minipage} \\ %\hline
{\Large{relajaito}} & {\Large{rilassa}} & {\Large{relaxte}} & \hspace{0.2cm}\begin{minipage}{.1\textwidth}
      \includegraphics[width=0.60\columnwidth,height=0.40cm]{farsi/chillax-3}
    \end{minipage} & {\Large{diooo}} & {\Large{oddiooo}} & {\Large{hahaha}} & \begin{minipage}{.1\textwidth}
      \includegraphics[width=0.60\columnwidth,height=0.40cm]{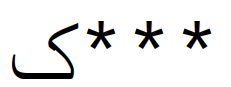}
    \end{minipage} \\ 
\bottomrule
\end{tabular}}
\caption{Translations of slang words and neologisms.% This illustrates how Twitter can be a valuable resource for creawting up-to-date bilingual dictionaries from scratch.
}
\label{tab:slang}
\end{table}

\subsection{Extrinsic evaluation: Cross-lingual sentiment analysis}
\label{sentanalysis}

In this section we test the performance of our cross-lingual embeddings in the sentiment analysis (SA) task%. SA is defined as the computational treatment of opinion, sentiment and subjectivity in text 
. We focus in particular on polarity classification \cite{pang_lee2008}.
\paragraph{Experimental setting}
%In order to perform the extrinsic evaluation, we selected a corpus of annotated tweets written in English as training data, and, as test data, tweets in Spanish and Italian.
%In order to perform the extrinsic evaluation, 
We selected an annotated dataset of English tweets as training data, and annotated datasets of Spanish, Italian and German tweets as test data. Since our main aim is the comparison of the cross-lingual embeddings, we used a standard Bidirectional Long Short-Term Memory (BiLSTM)  
architecture as classification system, with the same configuration across all experiments.\footnote{More details about the model and configuration (hyperparameters, etc.) are provided in the appendix.} % As classification system we used a standard Bidirectional Long Short-Term Memory (BiLSTM) %Recurrent Neural Network 
%architecture, with the same configuration across all experiments.\footnote{More details about the model and configuration (hyperparameters, etc.) are provided in the supplementary material.}
%In our experiments 
We used the cross-lingual embeddings for initializing the embedding layer. Given our cross-lingual evaluation setting, the weights of this embedding layer were not updated during training.  %three corpora of tweets written in Spanish, and a corpus of tweets written in Italian. 

\begin{table}[!tb]
    \footnotesize
    \centering
    %\begin{tabularx}{\linewidth}{@{}Xrrrr@{}}
    \begin{tabular*}{\linewidth}{@{\extracolsep{\fill}}lrrrr@{}}
    \toprule
    \textbf{Dataset}  &   \textbf{Positive}    &   \textbf{Neutral} &   \textbf{Negative}    &   \textbf{Total}\\
    \midrule
    SemEval\textsubscript{EN} &   3,094    &   2,043    &   863 &   5,999\\
    GCTASS\textsubscript{ES}  &   22,233    &   1,305 &   15,845 &   39,382\\
    InterTASS\textsubscript{ES}   &   642 &   216 &   768 &   1,625\\
    COST\textsubscript{ES}  &   5,637    &   -   &   5,789    &   11,426\\
    Sentipolc\textsubscript{IT}   &   316 &   255 &   734 &   1,305\\
    SB-10K\textsubscript{DE}      &   533 &   351 &   216 &  1,426\\

    \bottomrule
    \end{tabular*}
    \caption{%Size of the corpora used in the cross-SA evaluation.
    Size of the sentiment analysis datasets.}
    \label{tb_sa_corpus_size}
\end{table}

\newcolumntype{L}{>{\raggedright\arraybackslash}X}
\begin{table*}[!tb]
\begin{center}
\footnotesize
%\scalebox{0.7}
%\small
%\renewcommand{\arraystretch}{1.15}

%{
%\setlength{\tabcolsep}{4.0pt}

%\scalebox{0.9}{ 
\begin{tabularx}{\textwidth}{@{}L@{\hphantom{2}}Lr@{}r@{\hphantom{1}}r@{\hphantom{1}}r@{\hphantom{2}}r@{\hphantom{1}}r@{\hphantom{2}}r@{\hphantom{1}}r@{\hphantom{2}}r@{\hphantom{1}}r@{\hphantom{2}}r@{\hphantom{1}}r@{\hphantom{2}}r@{\hphantom{1}}r@{\hphantom{2}}r@{\hphantom{1}}r@{\hphantom{2}}r@{\hphantom{1}}r@{}}
%\cline{2-5}
\toprule
%&&	\multicolumn{10}{c}{\sc Spanish}	& \multicolumn{4}{c}{\sc Italian}	& \multicolumn{4}{c}{\sc German}\\ 
%\cmidrule(r){3-12}
%\cmidrule{13-16}
%\cmidrule(l){17-20}
\multirow{3}{*}{\textbf{\shortstack{Super-\\vision}}} 	& \multirow{3}{*}{\textbf{Model}} &   \multicolumn{2}{c}{\textbf{COST (ES)}} &  \multicolumn{4}{c}{\textbf{GCTASS (ES)}} &  \multicolumn{4}{c}{\textbf{InterTASS (ES)}} &  \multicolumn{4}{c}{\textbf{Sentipolc (IT)}} &  \multicolumn{4}{c}{\textbf{SB-10K (DE)}}\\ 	
%\cmidrule(r){3-12}
%\cmidrule(l){13-16}
%& & \multicolumn{2}{c}{\multirow{2}{*}{\textbf{COST}}} &  \multicolumn{4}{c}{\textbf{GCTASS}} &  \multicolumn{4}{c}{\textbf{InterTASS}} &  \multicolumn{4}{c}{\textbf{Sentipolc}} \\
\cmidrule(r){3-12}
\cmidrule(r){13-16}
\cmidrule{17-20}
&   &  \multicolumn{2}{c}{\textbf{2-class}}  &   \multicolumn{2}{c}{\textbf{3-class}}    &   \multicolumn{2}{c}{\textbf{2-class}}    &   \multicolumn{2}{c}{\textbf{3-class}}    &   \multicolumn{2}{c}{\textbf{2-class}}    &   \multicolumn{2}{c}{\textbf{3-class}}    &   \multicolumn{2}{c}{\textbf{2-class}}    &   \multicolumn{2}{c}{\textbf{3-class}}    &   \multicolumn{2}{c}{\textbf{2-class}}\\
\cmidrule(r){3-12}
\cmidrule(r){13-16}
\cmidrule{17-20}
& & \textbf{F1} &  \textbf{Acc} &  \textbf{F1} &  \textbf{Acc} & \textbf{F1} &  \textbf{Acc} & \textbf{F1} &  \textbf{Acc} & \textbf{F1} &  \textbf{Acc} & \textbf{F1} &  \textbf{Acc}& \textbf{F1} &  \textbf{Acc} & \textbf{F1} &  \textbf{Acc}& \textbf{F1} &  \textbf{Acc}\\
\cmidrule(r){1-12}
\cmidrule(r){13-16}
\cmidrule{17-20}
\multirow{2}{*}{\shortstack{Unsu-\\pervised}}   	& MUSE	&  50.0	& 58.0 & 5.2 &  26.1	& 37.0 & 58.4 &  24.0 &  28.2  &  32.0 &  45.8  &  22.1 &  25.9  &  25.3 &  31.1 &   27.8   &   33.4   &   \textbf{60.0}   &   62.5\\
  	&   VecMap	&  57.9	& 61.5 & 22.8 &  35.6	& 37.8 & 57.6 &  22.9 &  27.4 &  33.7 &  45.3  &  21.2 &  24.5  &  26.3 &  32.0 &   36.6   &   45.8   &   56.1   &   56.2\\
\cmidrule(r){1-12}
\cmidrule(r){13-16}
\cmidrule{17-20}
%Numerals &   \multirow{2}{*}{Artetxe et al (2017)}	& 47.9	& 51.3 & - 	\\
%Emoji  &  	& 47.3	& 51.7 & -     \\
%\hline
\multirow{5}{*}{Distant}   &   MUSE	&  49.4	& 57.7 & 25.1 &  52.3	& 37.9 & 58.3 &  22.9 &  33.8 &  35.1 &  45.9 &  23.3 &  25.4 &  28.1 &  32.8  &   41.7   &   50.8   &   56.8   &   57.1\\
   &  VecMap &  46.8	& 56.2 & 24.1 &  43.4	& 37.0 & 58.3 &  25.5 &  36.4 &  33.1 &  46.2 &  22.3 &  25.7 &  27.0 &  32.3  &   40.0   &   50.3   &   55.3   &   60.0\\
   &   Meemi	&  45.1	& 54.9 & 25.5 &  41.0	& 37.2 & 58.0 &  26.3 &  36.9 &  33.0 &  45.1 &  24.8 &  25.7 &  25.0 &  31.0  &   41.4   &   58.6   &   58.1   &   58.6\\
   &   \textit{Plain}	&  77.4	& 77.5 & 33.0 &  46.2	& 50.7 & 62.4 &  33.4 &  33.4 &  63.4 &  63.4 &  26.7 &  28.7 &  36.7 &  38.1 &   42.2   &   47.5   &   57.9   &   63.8\\
   &   \textit{Weighted}	&  \textbf{80.4}	& \textbf{80.5} & \textbf{42.6} &  \textbf{53.5}	& \textbf{64.7} & \textbf{66.2} &  \textbf{45.3} &  \textbf{51.7} &  \textbf{65.9} &  \textbf{67.2} &  \textbf{30.7} &  \textbf{32.0} &  \textbf{51.3} &  \textbf{51.5}    &   \textbf{44.8}    &   57.3   &   57.7   &   \textbf{65.4}\\
   %\midrule
   \cmidrule(r){1-12}
    \cmidrule(r){13-16}
    \cmidrule{17-20}
    
    \multirow{2}{*}{\shortstack{100\\pairs}}  & VecMap	&  58.0	& 63.1 & 25.8 & 50.6	& 37.1 & 58.3 &  24.2 &  39.5  &  32.0 &  45.8  &  24.3 &  27.2  &  25.6 &  31.3 &  36.1   &   56.6    &   57.2   &   62.3 \\
    &  MUSE	&  56.8	& 62.4 & 26.0 & 43.9	& 37.6 & 58.4 &  23.7 &  35.4  &  32.1 &  45.9  &  23.6 &  25.1  &  23.9 &  30.4 & 42.0 &  \textbf{60.4}    &   56.8   &   63.3 \\
    \cmidrule(r){1-12}
    \cmidrule(r){13-16}
    \cmidrule{17-20}
   \multirow{2}{*}{\textit{\shortstack{Lower\\bounds}}}  & Majority	&  33.0	& 49.3 & 24.1 &  56.5	& 36.9 & 58.4 &  18.9 &  39.5  &  31.3 &  45.6  &  13.0 &  24.2  &  23.1 &  30.1 & 12.4 &   23.0    &   37.6   &   60.2\\
 % 	&   Random	&  -	& - & - &  -	& - & - &  - &  - &  - &  -  &  - &  -  &  - &  - \\
  	  	&   FT~(EN)	&  50.2	& 57.1 & 23.2 &  36.0	& 37.5 & 58.0 &  21.9 &  37.2 &  32.3 &  43.9  &  25.6 &  26.3  &  36.6 &  37.6  &   40.2   &   55.2   &   58.0   & 58.0\\
  	%  	&   FastText (EN-ES/IT)	&  57.78	& 61.35 & 25.95 &  46.13	& 40.68 & 58.41 &  23.27 &  34.15 &  34.14 &  44.22  &  22.39 &  25.90  &  34.28 &  36.00 \\
\cmidrule(r){1-12}
\cmidrule(r){13-16}
\cmidrule{17-20}
%\midrule
\textit{Upper bound}	&   FT (ES/IT/DE)	&  87.9	& 87.9 & 56.1 &  78.4	& 80.5 & 81.0 &  49.7 &  59.2  &  71.1 &  71.8  &  49.4 &  53.6  &  73.3 &  75.8  &   62.1   &   72.3   &   76.1 &   77.4\\

\bottomrule
\end{tabularx}
\end{center}
\caption{\label{tab:sentan} Macro-average F1 and accuracy (\%) results in the cross-lingual SA evaluation, using different embeddings as features.  The \textit{Plain} and \textit{Weighted} models are built with our proposed post-processing technique over the base VecMap model. %The models Plain and Weighted are proposed in this paper.
}
\end{table*}

%\smallskip
\paragraph{Datasets}
%\begin{description}[style=unboxed,leftmargin=0pt]
%    \item[dd] d
%\end{description}
As training data we used %selected a corpus of tweets written in English, specifically 
the English dataset of the Sentiment Analysis in Twitter task of SemEval 2016 \cite{preslav2016}. %The tweets are annotated as being %on three levels of opinion intensity, namely 
%\texttt{Pos.}, \texttt{neutral} or \texttt{Neg.}. 
For evaluation we used %several datasets, %three of them are composed of tweets written in Spanish and one in Italian, 
the General Corpus of TASS (GCTASS) \cite{villenaroman2013}, COST \cite{martinezcamara2015} and InterTASS \cite{diazgaliano2018} for Spanish, Sentipolc \cite{barbieri2016evalita} for Italian and SB-10K \cite{cieliebak2017} for German. Table \ref{tb_sa_corpus_size} lists statistics of these datasets. We carried out both two-class (positive and negative) %(\texttt{Pos.} and \texttt{Neg.}) 
and three-class (positive, neutral and negative) evaluations with the %SemEval, 
GCTASS, InterTASS, Sentipolc and SB-10K datasets%\todo{by removing the neutral tweets in the two-class setting?}
, and a two-class evaluation with COST.

%The GCTASS corpus \cite{villenaroman2013} is formed of tweets written in Spanish, and it was the main corpus used in the workshop TASS from the edition of 2012 to the edition of 2016, it thus widely known by the Spanish research community. The corpus is annotated on six and three levels of opinion intensity, and we used the three point-scale version.

%\footnote{\url{http://www.sepln.org/workshops/tass/}}

%The InterTASS corpus \cite{diazgaliano2018} is composed of tweets written in Spanish, and it is the main corpus of the workshop TASS since the edition of 2017. As the GCTASS corpus, it is annotated on three levels of opinion intensity. The third corpus of tweets written in Spanish is COST \cite{martinezcamara2015}, in which the tweets are only positive or negative, and it was annotated following a distant supervision approach. The corpus of Italian tweets is Sentipolc \cite{basile2016}, whose tweets are annotated on three levels of opinion intensity. Table \ref{tb_sa_corpus_size} shows the size of the datasets.

%\subsubsection{Experimental setting}

%\smallskip
\paragraph{Lower and upper bounds} In addition to the comparison systems, in this experiment we also considered two lower bound systems and one upper bound, aimed at providing a broader context for our experimental results. %for analyzing the performance of the different set of embeddings. 
As lower bound systems we included: \begin{enumerate*}[label={(\arabic*)}] 
%\item a majority class classification system according to the training corpus (SemEval); 
\item always predicting the majority class from the SemEval 2016 training corpus; 
and \item training and testing the neural network with a set of monolingual English embeddings (FastText EN).\end{enumerate*} This latter baseline is introduced as a sanity check, as its only source for cross-lingual transfer comes from the fact that the vocabularies of different languages may overlap. The upper bound is a monolingual BiLSTM classification system which is trained for each test dataset using the associated training data.  %classification system. %vanilla neural network.

%We carried out both two-class and three-class classification evaluations with the SemEval, GCTASS, InterTASS and Sentipolc datasets, and a two-class evaluation with COST corpus. %However, we also performed the two-class evaluation with the corpora GCTASS, InterTASS and Sentipolc.

%For the sake of setting a lower bound results, we developed three baselines: \begin{enumerate*}[label={(\arabic*)}] \item a random classification system; \item a majority class classification system; \item training and test the neural network with a set of monolingual English embeddings (FastText EN); and \item training the system with a set of monolingual English embeddings, and test with a monolingual set of embeddings of each of the evaluation languages, Spanish and Italian (FasText EN-ES/IT). \end{enumerate*} We used the macro-average F1 score and the Accuracy as evaluation measures.

%Before performing the cross-embedding evaluation, we set the hyperparameters of our vanilla neural network. First, we developed a three-class supervised classification based on a bag-of-words representation weighted with TF-IDF, and SVM as classification algorithm. The SVM baseline system reached 45.30\% of F1 and 52.99\% of Accuracy, which may be ranked as the 28\textsuperscript{th} position at the SemEval 2016 competition \cite{preslav2016}. We tuned the neural network to outperform the F1 score reached by the SVM baseline, and the results reached are 47.34\% of F1 and 50.01\% of Accuracy, which may be ranked in the position 27\textsuperscript{th} at the SemEval 2016 competition.

%\smallskip
\paragraph{Results}
Table \ref{tab:sentan} summarizes the results for this cross-lingual SA evaluation. %in terms of classification accuracy and macro-averaged F1. 
Our main findings, which are consistent for the three languages, are as follows: \begin{enumerate*}[label={(\alph*)}] \item there are no large differences between the unsupervised and distantly supervised variants of MUSE and VecMap, which in general behave similarly to the two lower bound baselines%, with the exception of the COST corpus, which may be motivated by the dates of the tweets of COST corpus, 2011, and the dates of tweets used to compiled the dictionaries
; %\item the state-of-the-art cross-lingual embeddings reached results close to the two lower bounds, hence they do not provide an useful representation for at least cross-SA in Twitter; 
\item the results of the Meemi post-processing technique are also in line with the base VecMap model; \item our two post-processing techniques 
%prove beneficial for mapping further integrating the monolingual spaces, as the results are clearly superior to
lead to substantial improvements over the base VecMap model, including its weakly-supervised variant; %, and in some cases even approach the upper bound; 
and \item using frequency weighting %the average by the frequency (Ours weighted) 
clearly outperforms the unweighted variant of our model, with peak performances on COST and InterTASS.
%corpus (over 80\% in accuracy) and the InterTASS corpus (only five points below the upper bound in terms of F1 in both classification settings).
\end{enumerate*} 

In general, the results provided by our simple post-processing technique are encouraging, especially taking into account that (1) these embeddings were learned without making use of any external resources or bilingual data, (2) no data in the target language was used for training, and (3) the distribution of the English dataset used for training clearly differs from all these datasets (see Table \ref{tb_sa_corpus_size}). What is particularly surprising is the performance gap of our proposed technique with respect to the state-of-the-art cross-lingual embeddings of VecMap and MUSE. In fact, our weighted postprocessing technique leads to improvements of over 40\% over the base models in most cases.
%Therefore, the extrinsic evaluation shows the sufficiency of our proposal to generate sets of cross-lingual embeddings that share information from two languages, and they are useful for the task of cross-SA in Twitter. 
%To have a better understanding, we now further analyze the cross-lingual embeddings, both quantitatively, by means of a more intrinsic evaluation, and qualitatively.

\newcolumntype{L}{>{\raggedright\arraybackslash}X}
\begin{table*}[!tb]
\begin{center}
\footnotesize
\begin{tabularx}
{\textwidth}{@{}LLr@{\hphantom{0}}r@{\hphantom{0}}r@{\hphantom{1}}r@{\hphantom{0}}rr@{\hphantom{0}}r@{\hphantom{0}}r@{\hphantom{1}}r@{\hphantom{0}}rr@{\hphantom{0}}r@{\hphantom{0}}r@{\hphantom{1}}r@{\hphantom{0}}rrr@{\hphantom{0}}r@{}}
%{\textwidth}{@{}L@{\hphantom{0}}L@{\hphantom{2}}r@{\hphantom{1}}r@{\hphantom{1}}r@{\hphantom{1}}r@{\hphantom{1}}r@{\hphantom{1}}r@{\hphantom{2}}r@{\hphantom{1}}r@{\hphantom{1}}r@{\hphantom{1}}r@{\hphantom{1}}r@{\hphantom{1}}r@{\hphantom{2}}r@{\hphantom{1}}r@{\hphantom{1}}r@{\hphantom{2}}r@{\hphantom{1}}r@{\hphantom{1}}r@{\hphantom{1}}r@{}}
\toprule
\multirow{3}{*}{\bf Model} 	& \multirow{3}{*}{\textbf{\shortstack{Super-\\vision}}} 	&	\multicolumn{5}{c}{\bf EN-ES}	& \multicolumn{5}{c}{\bf EN-IT} & \multicolumn{5}{c}{\bf EN-DE}	& \multicolumn{3}{c}{\bf EN-FA}	\\ 
\cmidrule(lr){3-7}
\cmidrule(r){8-12}
\cmidrule(r){13-17}
\cmidrule{18-20}
    & & \multicolumn{3}{c}{\textbf{Word trans.}} &  \multicolumn{2}{c}{\textbf{SA}} &  \multicolumn{3}{c}{\textbf{Word trans.}} & \multicolumn{2}{c}{\textbf{SA}}  & \multicolumn{3}{c}{\textbf{Word trans.}} & \multicolumn{2}{c}{\textbf{SA}}  & \multicolumn{3}{c}{\textbf{Word trans.}}  \\
\cmidrule(lr){3-7}
\cmidrule(r){8-12}
\cmidrule(r){13-17}
\cmidrule{18-20}
& & P1 & P5  &  P10  & F1  & Acc & P1 & P5  & P10  & F1  & Acc & P1 & P5  & P10  & F1  & Acc &  P1 & P5 & P10\\
%\cmidrule(l{4pt}r{4pt}){2-4} \cmidrule(l{4pt}r{4pt}){5-7} 
%\multicolumn{1}{c|}{} &   {\bf F-Measure}  & {\bf F-Measure}   \\ 
%\cmidrule(l{4pt}r{4pt}){1-1} \cmidrule(l{4pt}r{4pt}){2-4} \cmidrule(l{4pt}r{4pt}){5-7} 

%\hline
%SW2V(Wiki) & - &  -	 \\
\cmidrule(r){1-7}
\cmidrule(r){8-12}
\cmidrule(r){13-17}
\cmidrule{18-20}
\multirow{5}{*}{VecMap}   	& \textit{All}	&  \bf 8.5	& 16.9 & \bf 21.6	& 25.5 & \textbf{36.4} &  \bf 9.1 &  16.8 &  21.8 & 22.3 &  \textbf{25.7}	& 2.6 & \bf 6.7 & \bf  9.6 & 40.0 & 50.3 &  0.2  &  0.5 &  1.1 \\
 % 	&  \hspace{0.2cm} - Numerals	&  -	& - & - &  -	& - & - &  - &  - &  -	& - & - &  -	& - & - &  - &  -  &  - &  - &  -     \\
  %	  	&   \hspace{0.2cm} - Emoji	&  -	& - & - &  -	& - & - &  - &  - &  -	& - & - &  -	& - & - &  - &  -  &  - &  - &  -     \\
  %	  	  	&  \hspace{0.2cm} - Others	&  -	& - & - &  -	& - & - &  - &  - &  -	& - & - &  -	& - & - &  - &  -  &  - &  - &  -     \\
  %	  	  	\cline{2-21}
  	  	  	  	&   Numerals	&  7.6	& 15.7 & 20.2	& 23.1 & 36.1 &  8.6 &  17.2 &  21.9 & \bf 23.6 &  24.8	& 2.7 & 6.4 & 9.3 & 38.9 & 43.4 &  0.0  &  0.0 &  0.0      \\
  	  	  	  	  	&  Emoji	&  7.8	& 16.9 & 21.2 	& \textbf{27.3} & 32.6 &  8.6 &  16.8 &  21.8 & 22.6  &  24.7	& \bf 3.1 & 6.2 &  8.5 & 44.3 & 55.5 &  \bf 0.5  &  1.3 &  1.7   \\
  	  	  	  	  	  	&  Words	&  8.1	& \bf 17.0 & \bf 21.6 	& 23.5 & 31.6 &  8.8 &  \bf 17.5 &  22.0 & 20.7 &  24.6	& 2.8 & 6.5 &  8.7 & \bf 44.8 & \bf 57.7 &  0.3  &  \bf 1.4 &  \bf 2.0     \\
  	  	  	  	  	  	%	\cline{2-21}
  	  	  	  	  	  		&  \textit{Unsup.}	&  8.1	& 16.4 & 20.4	& 22.9 & 27.4 &  8.8 &  17.0 &  \bf 22.3	 & 21.2 &  24.5	&  0.1 & 0.4 &  0.5 & 36.6 & 45.8 &  0.0  &  0.0 &  0.0    \\
%\midrule
\cmidrule(r){1-7}
\cmidrule(r){8-12}
\cmidrule(r){13-17}
\cmidrule{18-20}

%Numerals &   \multirow{2}{*}{Artetxe et al (2017)}	& 47.9	& 51.3 & - 	\\
%Emoji  &  	& 47.3	& 51.7 & -     \\
%\hline
%\multirow{4}{1em}{\shortstack{Weighted}} 
\multirow{4}{*}{\textit{Weighted}}   	& \textit{All}	& \bf 16.7 	& \bf 22.8 & \bf 28.5 	& \textbf{45.3} & \textbf{51.7} &  \bf 21.2 &  \bf 28.9 & \bf 33.5 & \textbf{30.7} &  \textbf{32.0}	& \bf 16.2 & \bf 18.0 & \bf 19.9 & \bf 44.8 &  57.3 &  \bf 1.2  &  \bf 1.5 &  \bf 1.8 \\
  %	&  \hspace{0.2cm} - Numerals	&  -	& - & - &  -	& - & - &  - &  - &  -	& - & - &  -	& - & - &  - &  -  &  - &  - &  -     \\
  	 % 	&  \hspace{0.2cm} - Emoji	&  -	& - & - &  -	& - & - &  - &  - &  -	& - & - &  -	& - & - &  - &  -  &  - &  - &  -     \\
  	  %	  	&  \hspace{0.2cm} - Others	&  -	& - & - &  -	& - & - &  - &  - &  -	& - & - &  -	& - & - &  - &  -  &  - &  - &  -     \\
  	  %	  	\cline{2-21}
  	  	  	  	&   Numerals	&  7.6	& 15.7 & 20.2 	& 23.0 & 25.4 & 8.5  &  17.1 &  21.9	 & 23.7 &  24.8	& 2.7 & 6.4 & 9.3 & 35.9 &  43.9 &  0.0  &  0.0 &  0.0    \\
  	  	  	  	  	&  Emoji	&  7.7	& 16.9 & 21.2 	& 24.5 & 34.3 &  8.5 &  16.8 &  21.8	 & 22.4 &  24.0	& 3.1 & 6.1 & 8.5 & 36.7 &  43.7 &  0.5  &  1.3 &  1.7     \\
  	  	  	  	  	  	&  Words	&  \bf 16.7	& 22.7 & 28.3 	& 37.2 & 37.4 &  \bf 21.2 & \bf 28.9 &  33.4	 & 21.6 &  25.4	& \bf 16.2 & 17.8 & 19.8 & \bf 44.8 &  \bf 57.7 &  \bf 1.2  &  1.4 &  \bf 1.8   \\
  	  	  	  	  	  	
  	  	  	  	  	  	%	&  Unsupervised	&  -	& - & - &  -	& - & - &  - &  - &  -	& - & - &  -	& - & - &  - &  -  &  - &  - &  - \\
\bottomrule
\end{tabularx}

\end{center}
\caption{\label{tab:ablation} Ablation test. Tasks: word translation (Word trans.) and cross-lingual sentiment analysis (SA).
}
\end{table*}

\paragraph{Analysis}

The main difference of our proposed averaging methods compared to VecMap and MUSE lies in the fact that they are creating \textit{anchor} points between languages. This turns out to be essential in a zero-shot cross-lingual transfer task. As argued in Section \ref{signals}, identical tokens such as emoji, numerals or homographs provide a reliable bilingual signal, and anchoring them to a middle point in the vector space facilitates the learning process. For example, the following Spanish tweet \textit{Buenos Dias a todos, menos a mi :(}  (\textit{Good Morning everyone, except for me}) was tagged as positive for both VecMap and MUSE, irrespective of their supervision. Similarly, VecMap and MUSE tagged the Italian tweet \textit{Alla ricerca del nirvana %\includegraphics[height=0.32cm,width=0.32cm]{emo/Smiling_Face_Emoji_large.png}
} (\textit{Looking for nirvana}) as negative. These systems thus overlooked a key emotion feature, i.e., :(, %and \includegraphics[height=0.32cm,width=0.32cm]{emo/Smiling_Face_Emoji_large.png}, 
and a critical loanword, i.e., \textit{nirvana}. %in sentiment analysis such as emojis and emoticons \cite{read2005}, i.e. :( and \includegraphics[height=0.32cm,width=0.32cm]{emo/Smiling_Face_Emoji_large.png}, and another key loanword like \textit{nirvana} in the Italian case. 
In contrast, the same sentiment analysis model trained with our \textit{weighted} cross-lingual embeddings correctly classified these two examples. 

\section{Ablation analysis}
\label{ablation}

As shown throughout all the experiments, using identical tokens as supervision proved more robust than fully-unsupervised methods. %In this section we analyze which types of identical words (see Section \ref{signals}) have a higher influence in the results
In order to get more insights from the results achieved in both evaluation tasks, we performed an ablation test on the different types of identical tokens in the synthetic dictionaries (see Section \ref{signals}). For this analysis, we focus on the base VecMap model and our proposed \textit{weighted} post-processing strategy.

Table \ref{tab:ablation} shows the results of this ablation test on the two considered tasks: word translation and cross-lingual sentiment analysis (SA).\footnote{Due to space constraints and for the sake of clarity, for this ablation test, Table \ref{tab:ablation} shows the results on the Facebook datasets on word translation and on the 3-class configuration on SA, using the most recent InterTASS dataset for Spanish.} %(the main conclusions are general for the other datasets as well).}
%\subsection{Ablation test}
%As shown throughout all the experiments, using identical words as supervision proved more robust than fully-unsupervised methods. In this section we analyze which types of identical words (see Section \ref{signals}) have a higher influence in the results, and if certain kinds of identical words can be removed to achieve an even superior performance...
%
Unsurprisingly, the dictionaries of shared words (i.e. identical tokens that are neither numerals nor emoji) provide the best results among the individual features, often being close to the full dictionary of identical tokens. This type of dictionary is the largest in size, comprising over 95\% of all identical tokens in the cases of Spanish and Italian. However, for the base VecMap model dictionaries consisting of either numerals or emoji seem to be enough to achieve similar results in most tasks and languages. %In fact, using only emojis as distant supervision in SA achieves similar or better performance than using all remaining shared words in Spanish and Italian. These results show the relevance of the information underlying in emojis, which have been proved helpful on their own in tasks such as sentiment analysis \cite{read2005}.
%\stevenNote{I don't think we need the following sentence (and if we keep it, it needs to be improved as it is not very clear at the moment).}
This is not the case when using our weighted postprocessing, which highlights its potential for taking advantage of the heterogeneity of all identical tokens. In fact, using the dictionary of all identical tokens consistently provides the best results in all tasks and languages (including Farsi) except in one single measure in German SA. %Interestingly, the cross-lingual embeddings exploiting only emoji obtain good overall results for Farsi. %Japanese, and on similarity for Farsi. 

\section{Conclusion}
%\stevenNote{Here we again have this ambiguity about what the nature of the contribution is. The conclusion is presented as if the postprocessing strategy is only a side-result and the main contribution is the comprehensive evaluation of existing methods on social media.}
The main contribution of this paper is two-fold. On the one hand, we have presented a comprehensive study on the performance of state-of-the-art methods for learning cross-lingual embeddings without external bilingual data in the domain of social media communication. % by means of monolingual corpora only, without any external resources or human supervision. 
The overall results are encouraging, as they show that high-quality cross-lingual embeddings can be obtained directly from noisy user-generated corpora without external resources via distant supervision. These embeddings can be leveraged for cross-lingual downstream applications where training data may be scarce, as shown in our sentiment analysis experiments. However, our evaluation suggests there is significant room for improvement overall. Our results show that, especially in the case of distant languages such as English-Farsi, state-of-the-art cross-lingual mappings fail to learn an accurate mapping between the languages. %, especially in the fully unsupervised setting. %which do not share many properties with the English language, such as a similar topology or the usage of the same alphabet. %apart from the minimum necessary tools and raw training text corpora.

%In general the evaluation clearly showed that methods exploiting implicit bilingual signals such as identical words are more robust than other methods which construct the bilingual signals from the distribution of the monolingual embeddings. 
On the other hand, we have also introduced a simple post-processing technique which alters the embeddings of tokens that appear in both languages by simply averaging their initial embeddings.
%by exploiting a weighted average through a synthetic bilingual dictionary which can be automatically obtained from monolingual corpora. 
%The results show that d
Despite its simplicity, our proposed technique clearly improves the quality of state-of-the-art cross-lingual word embedding approaches. %With the inclusion of a simple weighting scheme % in the case of closely-related languages. %in both intrinsic and extrinsic tasks. 
In fact, we showed how a standard sentiment analysis system can achieve results of up to 80\% in accuracy without the need of any training data in the test language by using our proposed method, improving the state of the art by more than 40\% in several cases. The results also suggest that our method can be further improved by tuning it to specific applications or by exploiting the underlying idea to local neighbours in the vector space, amplifying its impact.
% intrinsic and extrinsic tasks
% ... and semi-supervised is better than unsupervised?
In general, the simplicity of our approach opens up exciting avenues of research on cross-lingual applications in social media, where annotated data in English %(or any language for which data in a particular task is available) 
can be exploited for other languages with few resources.  %Finally, the simplicity of our approach opens up exciting avenues for future research. For example, we believe that a more carefully tuned model exploiting the signals proposed in this paper could help learn cross-lingual embeddings of an even higher quality. %Another obvious direction is to 
The construction of cross-lingual embedding models also paves the way for the development of unsupervised machine translation systems %, as shown in previous works 
\cite{artetxe2018iclr,lample:2018emnlp}, %, which are based on the unsupervised cross-lingual embedding of \citeauthor{artetxe-labaka-agirre:2017:Long} \shortcite{artetxe-labaka-agirre:2017:Long} and \citeauthor{conneau2018word} \shortcite{conneau2018word}. 
in this case %they would %following the line of work started by \cite{artetxe2018iclr} and \cite{lample2018unsupervised},
specifically targeting noisy user-generated text for which parallel data is extremely scarce, and not even available at all for widely spoken language pairs. Indeed, standard machine translation tools are generally %based on large parallel corpora, as they 
not suited for the kind of noisy text that is found in social media, where the language used is very dynamic and new terms are constantly being introduced. %Moreover, the reliance on solely monolingual corpora which can be easily obtained introduces the possibility of improving and updating the cross-lingual embeddings at almost live time.

%\end{spacing}
\section*{Acknowledgements}
Steven Schockaert was supported by ERC Starting Grant 637277. Yerai Doval is supported by the Spanish Ministry of Economy, Industry and Competitiveness (MINECO) through projects FFI2014-51978-C2-2-R, TIN2017–85160–C2–1-R and TIN2017–85160–C2–2-R; the Spanish State Secretariat for Research, Development and Innovation and the European Social Fund (ESF) under an FPI fellowship (BES-2015-073768) associated to project FFI2014-51978-C2-1-R; and by the Galician Regional Government under project ED431D 2017/12. Eugenio Mart\'{i}nez C\'{a}mara was supported by the Spanish Government Programme Juan de la Cierva Formaci\'{o}n (FJCI-2016-28353). Finally, we would like to thank Taher Pilehvar for his help with the Farsi qualitative analysis.

\bibliography{aaai20}

\begin{thebibliography}{}

\bibitem[\protect\citeauthoryear{Artetxe \bgroup et al\mbox.\egroup
  }{2018}]{artetxe2018iclr}
Artetxe, M.; Labaka, G.; Agirre, E.; and Cho, K.
\newblock 2018.
\newblock Unsupervised neural machine translation.
\newblock In {\em Proceedings of ICLR}.

\bibitem[\protect\citeauthoryear{Artetxe, Labaka, and
  Agirre}{2017}]{artetxe-labaka-agirre:2017:Long}
Artetxe, M.; Labaka, G.; and Agirre, E.
\newblock 2017.
\newblock Learning bilingual word embeddings with (almost) no bilingual data.
\newblock In {\em Proceedings of ACL},  451--462.

\bibitem[\protect\citeauthoryear{Artetxe, Labaka, and
  Agirre}{2018a}]{artetxe2018generalizing}
Artetxe, M.; Labaka, G.; and Agirre, E.
\newblock 2018a.
\newblock Generalizing and improving bilingual word embedding mappings with a
  multi-step framework of linear transformations.
\newblock In {\em Proceedings of AAAI}.

\bibitem[\protect\citeauthoryear{Artetxe, Labaka, and
  Agirre}{2018b}]{artetxe:acl2018}
Artetxe, M.; Labaka, G.; and Agirre, E.
\newblock 2018b.
\newblock A robust self-learning method for fully unsupervised cross-lingual
  mappings of word embeddings.
\newblock In {\em Proceedings of ACL},  789--798.

\bibitem[\protect\citeauthoryear{Barbieri \bgroup et al\mbox.\egroup
  }{2016a}]{barbieri2016evalita}
Barbieri, F.; Basile, V.; Croce, D.; Nissim, M.; Novielli, N.; and Patti, V.
\newblock 2016a.
\newblock Overview of the evalita 2016 sentiment polarity classification task.
\newblock In {\em Proceedings of CLiC-it}.

\bibitem[\protect\citeauthoryear{Barbieri \bgroup et al\mbox.\egroup
  }{2016b}]{barbieri2016cosmopolitan}
Barbieri, F.; Kruszewski, G.; Ronzano, F.; and Saggion, H.
\newblock 2016b.
\newblock How cosmopolitan are emojis?: Exploring emojis usage and meaning over
  different languages with distributional semantics.
\newblock In {\em Proceedings of ACMM},  531--535.

\bibitem[\protect\citeauthoryear{Barnes, Klinger, and Schulte~im
  Walde}{2018}]{barnes-etal-2018-bilingual}
Barnes, J.; Klinger, R.; and Schulte~im Walde, S.
\newblock 2018.
\newblock Bilingual sentiment embeddings: Joint projection of sentiment across
  languages.
\newblock In {\em Proceedings of the 56th Annual Meeting of the Association for
  Computational Linguistics (Volume 1: Long Papers)},  2483--2493.
\newblock Melbourne, Australia: Association for Computational Linguistics.

\bibitem[\protect\citeauthoryear{Bojanowski \bgroup et al\mbox.\egroup
  }{2017}]{bojanowski2017enriching}
Bojanowski, P.; Grave, E.; Joulin, A.; and Mikolov, T.
\newblock 2017.
\newblock Enriching word vectors with subword information.
\newblock {\em Transactions of the Association of Computational Linguistics}
  5(1):135--146.

\bibitem[\protect\citeauthoryear{Cappallo, Mensink, and
  Snoek}{2015}]{cappallo2015image2emoji}
Cappallo, S.; Mensink, T.; and Snoek, C.~G.
\newblock 2015.
\newblock Image2emoji: Zero-shot emoji prediction for visual media.
\newblock In {\em Proceedings of ACMM},  1311--1314.

\bibitem[\protect\citeauthoryear{Chen and Cardie}{2018}]{chen2018unsupervised}
Chen, X., and Cardie, C.
\newblock 2018.
\newblock Unsupervised multilingual word embeddings.
\newblock In {\em Proceedings of the 2018 Conference on Empirical Methods in
  Natural Language Processing},  261--270.

\bibitem[\protect\citeauthoryear{Chen \bgroup et al\mbox.\egroup
  }{2018}]{chen2018adversarial}
Chen, X.; Sun, Y.; Athiwaratkun, B.; Cardie, C.; and Weinberger, K.
\newblock 2018.
\newblock Adversarial deep averaging networks for cross-lingual sentiment
  classification.
\newblock {\em Transactions of the Association for Computational Linguistics}
  6:557--570.

\bibitem[\protect\citeauthoryear{Chollet}{2015}]{chollet2015}
Chollet, F.
\newblock 2015.
\newblock Keras.
\newblock \url{https://github.com/fchollet/keras}.

\bibitem[\protect\citeauthoryear{Cieliebak \bgroup et al\mbox.\egroup
  }{2017}]{cieliebak2017}
Cieliebak, M.; Deriu, J.~M.; Egger, D.; and Uzdilli, F.
\newblock 2017.
\newblock A twitter corpus and benchmark resources for german sentiment
  analysis.
\newblock In {\em Proceedings of the Fifth International Workshop on Natural
  Language Processing for Social Media},  45--51.
\newblock Association for Computational Linguistics.

\bibitem[\protect\citeauthoryear{Conneau \bgroup et al\mbox.\egroup
  }{2018}]{conneau2018word}
Conneau, A.; Lample, G.; Ranzato, M.; Denoyer, L.; and J{\'e}gou, H.
\newblock 2018.
\newblock Word translation without parallel data.
\newblock In {\em Proceedings of ICLR}.

\bibitem[\protect\citeauthoryear{Derczynski \bgroup et al\mbox.\egroup
  }{2013}]{derczynski2013twitter}
Derczynski, L.; Ritter, A.; Clark, S.; and Bontcheva, K.
\newblock 2013.
\newblock {T}witter part-of-speech tagging for all: Overcoming sparse and noisy
  data.
\newblock In {\em Proceedings of RANLP},  198--206.

\bibitem[\protect\citeauthoryear{D\'{i}az-Galiano \bgroup et al\mbox.\egroup
  }{2018}]{diazgaliano2018}
D\'{i}az-Galiano, M.~C.; Mart\'{i}nez-C\'{a}mara, E.; \'{A}ngel
  Garc\'{i}a~Cumbreras, M.; Vega, M.~G.; and Rom\'{a}n, J.~V.
\newblock 2018.
\newblock The democratization of deep learning in {TASS} 2017.
\newblock {\em Procesamiento del Lenguaje Natural} 60:37--44.

\bibitem[\protect\citeauthoryear{Dinu, Lazaridou, and
  Baroni}{2015}]{dinu2015improving}
Dinu, G.; Lazaridou, A.; and Baroni, M.
\newblock 2015.
\newblock Improving zero-shot learning by mitigating the hubness problem.
\newblock In {\em Proceedings of ICLR, Workshop track}.

\bibitem[\protect\citeauthoryear{Doval \bgroup et al\mbox.\egroup
  }{2018}]{doval:meemiemnlp2018}
Doval, Y.; Camacho-Collados, J.; Espinosa-Anke, L.; and Schockaert, S.
\newblock 2018.
\newblock Improving cross-lingual word embeddings by meeting in the middle.
\newblock In {\em Proceedings of EMNLP}.

\bibitem[\protect\citeauthoryear{Faruqui and Dyer}{2014}]{faruqui2014improving}
Faruqui, M., and Dyer, C.
\newblock 2014.
\newblock Improving vector space word representations using multilingual
  correlation.
\newblock In {\em Proceedings of EACL},  462--471.

\bibitem[\protect\citeauthoryear{Felbo \bgroup et al\mbox.\egroup
  }{2017}]{felbo2017using}
Felbo, B.; Mislove, A.; S{\o}gaard, A.; Rahwan, I.; and Lehmann, S.
\newblock 2017.
\newblock Using millions of emoji occurrences to learn any-domain
  representations for detecting sentiment, emotion and sarcasm.
\newblock In {\em Proceedings of EMNLP},  1615--1625.

\bibitem[\protect\citeauthoryear{Glava{\v{s}} \bgroup et al\mbox.\egroup
  }{2019}]{glavas-etal-2019-properly}
Glava{\v{s}}, G.; Litschko, R.; Ruder, S.; and Vuli{\'c}, I.
\newblock 2019.
\newblock How to (properly) evaluate cross-lingual word embeddings: On strong
  baselines, comparative analyses, and some misconceptions.
\newblock In {\em Proceedings of the 57th Annual Meeting of the Association for
  Computational Linguistics},  710--721.
\newblock Florence, Italy: Association for Computational Linguistics.

\bibitem[\protect\citeauthoryear{Godin \bgroup et al\mbox.\egroup
  }{2015}]{godin2015multimedia}
Godin, F.; Vandersmissen, B.; De~Neve, W.; and Van~de Walle, R.
\newblock 2015.
\newblock {Named Entity Recognition for Twitter Microposts using Distributed
  Word Representations}.
\newblock In {\em Proceedings of ACL Workshop on Noisy User-generated Text},
  146--153.

\bibitem[\protect\citeauthoryear{Hochreiter and
  Schmidhuber}{1997}]{hochreiter1997long}
Hochreiter, S., and Schmidhuber, J.
\newblock 1997.
\newblock Long short-term memory.
\newblock {\em Neural computation} 9(8):1735--1780.

\bibitem[\protect\citeauthoryear{Imran \bgroup et al\mbox.\egroup
  }{2018}]{imran2018processing}
Imran, M.; Castillo, C.; Diaz, F.; and Vieweg, S.
\newblock 2018.
\newblock Processing social media messages in mass emergency: Survey summary.
\newblock In {\em Companion Proceedings of the The Web Conference 2018},
  507--511.
\newblock International World Wide Web Conferences Steering Committee.

\bibitem[\protect\citeauthoryear{Lample \bgroup et al\mbox.\egroup
  }{2018}]{lample:2018emnlp}
Lample, G.; Ott, M.; Conneau, A.; Denoyer, L.; and Ranzato, M.
\newblock 2018.
\newblock Phrase-based {\&} neural unsupervised machine translation.
\newblock In {\em Proceedings of the 2018 Conference on Empirical Methods in
  Natural Language Processing},  5039--5049.

\bibitem[\protect\citeauthoryear{Lample, Denoyer, and
  Ranzato}{2018}]{lample2018unsupervised}
Lample, G.; Denoyer, L.; and Ranzato, M.
\newblock 2018.
\newblock Unsupervised machine translation using monolingual corpora only.
\newblock In {\em Proceedings of ICLR}.

\bibitem[\protect\citeauthoryear{Mart\'{i}nez-C\'{a}mara \bgroup et
  al\mbox.\egroup }{2014}]{martinezcamara2014}
Mart\'{i}nez-C\'{a}mara, E.; Mart\'{i}n-Valdivia, M.~T.; Ure\~{n}a L\'{o}pez,
  L.~A.; and Montejo-R\'{a}ez, A.
\newblock 2014.
\newblock Sentiment analysis in {T}witter.
\newblock {\em Natural Language Engineering} 20(1):1–28.

\bibitem[\protect\citeauthoryear{Mart\'{i}nez-C\'{a}mara \bgroup et
  al\mbox.\egroup }{2015}]{martinezcamara2015}
Mart\'{i}nez-C\'{a}mara, E.; Mart\'{i}n-Valdivia, M.~T.; Ureña-L\'{o}pez,
  L.~A.; and Mitkov, R.
\newblock 2015.
\newblock Polarity classification for {S}panish tweets using the {COST} corpus.
\newblock {\em Journal of Information Science} 41(3):263--272.

\bibitem[\protect\citeauthoryear{Meng \bgroup et al\mbox.\egroup
  }{2012}]{meng2012}
Meng, X.; Wei, F.; Liu, X.; Zhou, M.; Xu, G.; and Wang, H.
\newblock 2012.
\newblock Cross-lingual mixture model for sentiment classification.
\newblock In {\em Proceedings of the 50th Annual Meeting of the Association for
  Computational Linguistics (Volume 1: Long Papers)},  572--581.
\newblock Association for Computational Linguistics.

\bibitem[\protect\citeauthoryear{Mikolov \bgroup et al\mbox.\egroup
  }{2013}]{Mikolovetal:2013}
Mikolov, T.; Chen, K.; Corrado, G.; and Dean, J.
\newblock 2013.
\newblock Efficient estimation of word representations in vector space.
\newblock {\em CoRR} abs/1301.3781.

\bibitem[\protect\citeauthoryear{Mikolov, Le, and
  Sutskever}{2013}]{mikolov2013exploiting}
Mikolov, T.; Le, Q.~V.; and Sutskever, I.
\newblock 2013.
\newblock Exploiting similarities among languages for machine translation.
\newblock {\em arXiv preprint arXiv:1309.4168}.

\bibitem[\protect\citeauthoryear{Nair and Hinton}{2010}]{icml2010_NairH10}
Nair, V., and Hinton, G.~E.
\newblock 2010.
\newblock Rectified linear units improve restricted boltzmann machines.
\newblock In {\em Proceedings of the 27th International Conference on Machine
  Learning},  807--814.

\bibitem[\protect\citeauthoryear{Nakov \bgroup et al\mbox.\egroup
  }{2016}]{preslav2016}
Nakov, P.; Ritter, A.; Rosenthal, S.; Sebastiani, F.; and Stoyanov, V.
\newblock 2016.
\newblock Semeval-2016 task 4: Sentiment analysis in {T}witter.
\newblock In {\em Proceedings of SemEval},  1--18.

\bibitem[\protect\citeauthoryear{Pang and Lee}{2008}]{pang_lee2008}
Pang, B., and Lee, L.
\newblock 2008.
\newblock Opinion mining and sentiment analysis.
\newblock {\em Foundations and Trends in Information Retrieval} 2:1--135.

\bibitem[\protect\citeauthoryear{Pennington, Socher, and
  Manning}{2014}]{pennington2014glove}
Pennington, J.; Socher, R.; and Manning, C.~D.
\newblock 2014.
\newblock Glo{V}e: Global vectors for word representation.
\newblock In {\em Proceedings of EMNLP},  1532--1543.

\bibitem[\protect\citeauthoryear{Rasooli \bgroup et al\mbox.\egroup
  }{2018}]{Rasooli2018}
Rasooli, M.~S.; Farra, N.; Radeva, A.; Yu, T.; and McKeown, K.
\newblock 2018.
\newblock Cross-lingual sentiment transfer with limited resources.
\newblock {\em Machine Translation} 32(1):143--165.

\bibitem[\protect\citeauthoryear{Reyes, Rosso, and
  Buscaldi}{2012}]{reyes2012humor}
Reyes, A.; Rosso, P.; and Buscaldi, D.
\newblock 2012.
\newblock From humor recognition to irony detection: The figurative language of
  social media.
\newblock {\em Data \& Knowledge Engineering} 74:1--12.

\bibitem[\protect\citeauthoryear{Ruder, Vuli{\'c}, and
  S{\o}gaard}{2019}]{ruder2019survey}
Ruder, S.; Vuli{\'c}, I.; and S{\o}gaard, A.
\newblock 2019.
\newblock A survey of cross-lingual word embedding models.
\newblock {\em Journal of Artificial Intelligence Research} 65:569--631.

\bibitem[\protect\citeauthoryear{Salameh, Mohammad, and
  Kiritchenko}{2015}]{salameh2015}
Salameh, M.; Mohammad, S.; and Kiritchenko, S.
\newblock 2015.
\newblock Sentiment after translation: A case-study on arabic social media
  posts.
\newblock In {\em Proceedings of the 2015 Conference of the North American
  Chapter of the Association for Computational Linguistics: Human Language
  Technologies},  767--777.
\newblock Denver, Colorado: Association for Computational Linguistics.

\bibitem[\protect\citeauthoryear{Smith \bgroup et al\mbox.\egroup
  }{2017}]{smith2017offline}
Smith, S.~L.; Turban, D.~H.; Hamblin, S.; and Hammerla, N.~Y.
\newblock 2017.
\newblock Offline bilingual word vectors, orthogonal transformations and the
  inverted softmax.
\newblock In {\em Proceedings of ICLR}.

\bibitem[\protect\citeauthoryear{S{\o}gaard, Ruder, and
  Vuli{\'c}}{2018}]{sogaard2018limitations}
S{\o}gaard, A.; Ruder, S.; and Vuli{\'c}, I.
\newblock 2018.
\newblock On the limitations of unsupervised bilingual dictionary induction.
\newblock In {\em Proceedings of ACL},  778--788.

\bibitem[\protect\citeauthoryear{Tang \bgroup et al\mbox.\egroup
  }{2014}]{tang2014learning}
Tang, D.; Wei, F.; Yang, N.; Zhou, M.; Liu, T.; and Qin, B.
\newblock 2014.
\newblock Learning sentiment-specific word embedding for twitter sentiment
  classification.
\newblock In {\em Proceedings of ACL},  1555--1565.

\bibitem[\protect\citeauthoryear{Upadhyay \bgroup et al\mbox.\egroup
  }{2016}]{upadhyay2016cross}
Upadhyay, S.; Faruqui, M.; Dyer, C.; and Roth, D.
\newblock 2016.
\newblock Cross-lingual models of word embeddings: An empirical comparison.
\newblock In {\em Proceedings of the 54th Annual Meeting of the Association for
  Computational Linguistics (Volume 1: Long Papers)}, volume~1,  1661--1670.

\bibitem[\protect\citeauthoryear{Vilares, Alonso, and
  G\'{o}mez-Rodr\'{i}guez}{2017}]{vilares2017}
Vilares, D.; Alonso, M.~A.; and G\'{o}mez-Rodr\'{i}guez, C.
\newblock 2017.
\newblock Supervised sentiment analysis in multilingual environments.
\newblock {\em Information Processing \& Management} 53(3):595 -- 607.

\bibitem[\protect\citeauthoryear{Villena-Rom\'{a}n \bgroup et al\mbox.\egroup
  }{2013}]{villenaroman2013}
Villena-Rom\'{a}n, J.; Lana-Serrano, S.; Mart\'{i}nez-C\'{a}mara, E.; and
  Gonz\'{a}lez-Crist\'{o}bal, J.~C.
\newblock 2013.
\newblock {TASS} - workshop on sentiment analysis at sepln.
\newblock {\em Procesamiento del Lenguaje Natural} 50(0):37--44.

\bibitem[\protect\citeauthoryear{Vuli{\'c} and Korhonen}{2016}]{vulic2016role}
Vuli{\'c}, I., and Korhonen, A.
\newblock 2016.
\newblock On the role of seed lexicons in learning bilingual word embeddings.
\newblock In {\em Proceedings of the 54th Annual Meeting of the Association for
  Computational Linguistics (Volume 1: Long Papers)}, volume~1,  247--257.

\bibitem[\protect\citeauthoryear{Xing \bgroup et al\mbox.\egroup
  }{2015}]{xing2015normalized}
Xing, C.; Wang, D.; Liu, C.; and Lin, Y.
\newblock 2015.
\newblock Normalized word embedding and orthogonal transform for bilingual word
  translation.
\newblock In {\em Proceedings of NAACL},  1006--1011.

\bibitem[\protect\citeauthoryear{Xu \bgroup et al\mbox.\egroup
  }{2018}]{xu2018crosslingual}
Xu, R.; Yang, Y.; Otani, N.; and Wu, Y.
\newblock 2018.
\newblock Unsupervised cross-lingual transfer of word embedding spaces.
\newblock In {\em Proceedings of the 2018 Conference on Empirical Methods in
  Natural Language Processing},  2465--2474.
\newblock Association for Computational Linguistics.

\bibitem[\protect\citeauthoryear{Yang, Macdonald, and Ounis}{2018}]{Yang2018}
Yang, X.; Macdonald, C.; and Ounis, I.
\newblock 2018.
\newblock Using word embeddings in twitter election classification.
\newblock {\em Information Retrieval Journal} 21(2):183--207.

\bibitem[\protect\citeauthoryear{Zhang \bgroup et al\mbox.\egroup
  }{2017}]{zhangcrosslingual:emnlp2017}
Zhang, M.; Liu, Y.; Luan, H.; and Sun, M.
\newblock 2017.
\newblock Earth mover's distance minimization for unsupervised bilingual
  lexicon induction.
\newblock In {\em Proceedings of EMNLP},  1934--1945.

\bibitem[\protect\citeauthoryear{Zhou, Wan, and Xiao}{2016}]{zhou2016}
Zhou, X.; Wan, X.; and Xiao, J.
\newblock 2016.
\newblock Cross-lingual sentiment classification with bilingual document
  representation learning.
\newblock In {\em Proceedings of ACL},  1403--1412.

\end{thebibliography}
\bibliographystyle{aaai}

\section*{Appendix: Sentiment Analysis Classification System}

We provide specific details of the classification system used in our sentiment analysis experiments (Section 4.2 of the paper). As classification system, we made use of a standard Bidirectional Long Short-Term Memory (BiLSTM) recurrent neural network architecture \cite{hochreiter1997long}, developed in Keras \cite{chollet2015}. In the following we describe the details and hyperparameters of the specific architecture which is used across all experiments. The goal of the neural network is the classification of the opinion of tweets, hence the input is composed of a sequence of tokens of a tweet, and the output is the sentiment meaning of the tweet ($t$). %In order to tokenize the tweets, we used the Twitter-aware tokenizer of the Python library NLTK\footnote{\url{https://www.nltk.org/api/nltk.tokenize.html#module-nltk.tokenize.casual}}. Before the tokenization, we converted the tweets to lowercase, and we substituted all the Twitter user mentions (\textit{@username}) to the wild card expression \textit{@user}. The tokenization results in a sequence of tokens $t_{1:n}=\{t_{1}, \ldots, t_{n}\}$. We set the length of the input ($l$) to the median of the lengths of the training tweets in order to avoid a high amount of padding tokens. 
Specifically, the input of the neural network is the sequence of tokens $t_{1:l}$.

The first layer is the embeddings lookup layer, which returns the sequence $s\in {\rm I\!R}^{l\times 100}$. Since our aim is to test our cross-lingual embeddings, which were used to initialize the embedding layer, in a cross-lingual setting, the embedding weights are not updated during the training of the network. The output of the embedding layer is encoded by a %Bidirectional Long Short-Term Memory 
BiLSTM, which is an elaboration of two Long Short-Term Memory (LSTM) layers. One of the LSTM layers processes the sequence $s_{1:l}$ ($LSTM^{f}$), and the second one processes the sequence $s_{l:1}$ ($LSTM^{b}$). We concatenated the output of the two LSTM layers, which are the state vectors of each token of the sequence $s$. Since the number of internal units of each LSTM layer is 128, the output of the BiLSTM layer is the sequence $c \in {\rm I\!R}^{l\times 256}$.

%The developed neural network is composed of several layers. The first one is the input layer, which is the sequence of tokens of the tweets ($t$) whose length ($l$) was set to the median of the lengths of the training tweets. The subsequent layer is the embeddings lookup layer, which returns the sequence $s\in {\rm I\!R}^{l\times 100}$. Since our aim is to test our cross-lingual embeddings, which are used to initialize the embedding layer, in a cross-lingual setting, the embedding weights are not updated during the training of the network.

\begin{figure}[!t]
    \centering
    \includegraphics[width=0.65\linewidth, height=7.5cm]{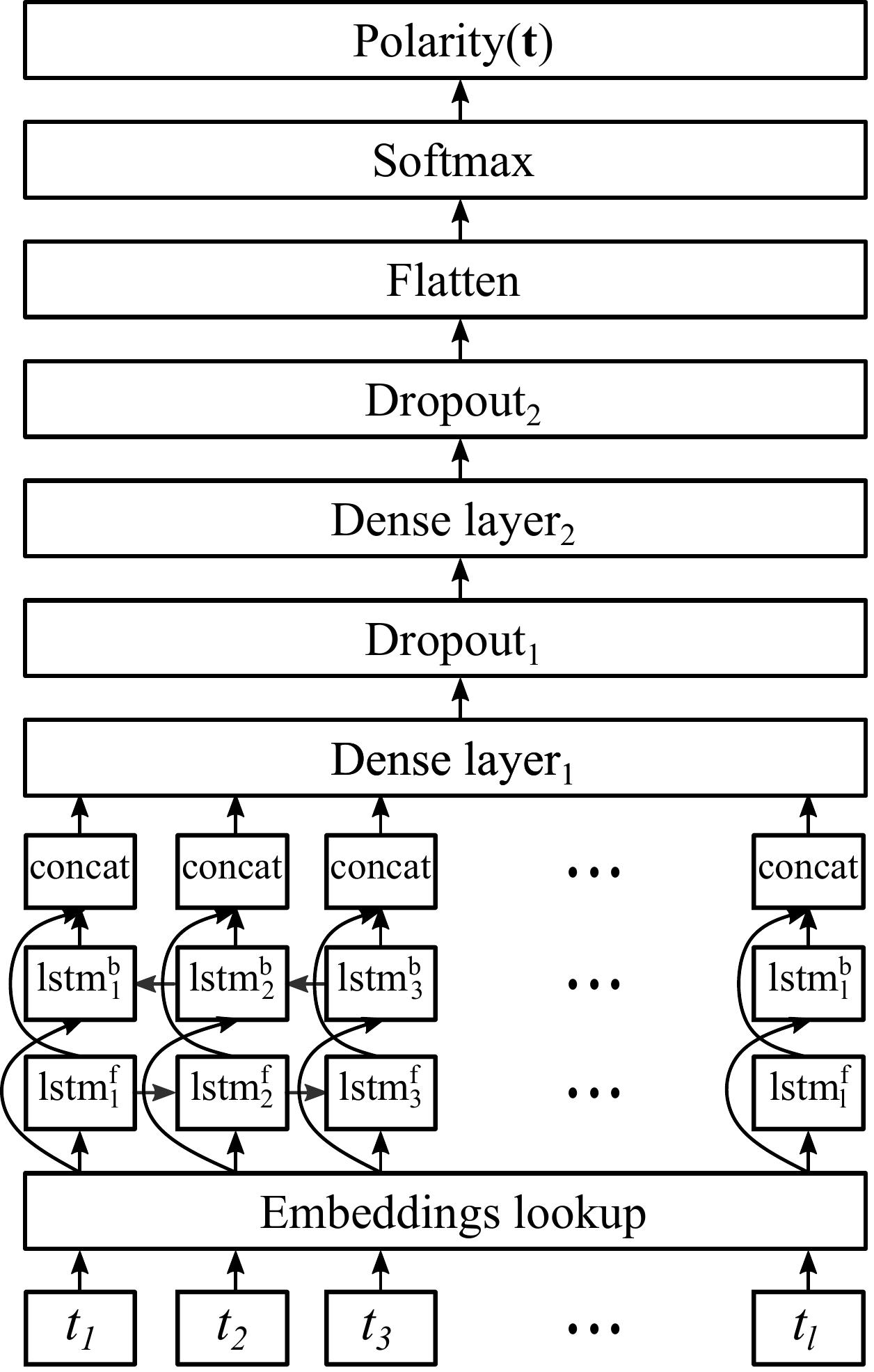}
    \caption{Architecture of the neural network developed for the cross-lingual sentiment analysis evaluation.}
    \label{fig_nn}
\end{figure}

%The first one is the input layer, which is the sequence of tokens of the tweets whose length ($l$) was set to the median of the lengths of the training tweets. {\color{red} Por que no todo el tweet? Ya que el limite de los tweets tampoco es muy grande} The subsequent layer is the embeddings lookup layer, which returns the sequence $s\in {\rm I\!R}^{l\times 100}$. %Since the aim is the evaluation of our cross-lingual embeddings, the value of the embeddings are not updated during the training of the network.
Two fully connected %or dense 
layers activated by the ReLU function \cite{icml2010_NairH10} process the sequence to the output of the BiLSTM layer. The output dimensions of the two fully connected layers are 64 and 32, respectively. A dropout layer is added after each fully connected layer, with a rate value of 0.5. $L_{2}$ regularization is applied to the weights of the fully connected layers with a value of 0.001, and to the output of the fully connected layers with a value of 0.0001. The output of the last fully connected layer is flattened, hence the dimension of the sequence $c$ after the processing of the two dense layers is ${\rm I\!R}^{l\times 32}$. The last layer is a softmax classification function. The output dimension of the softmax layer depends on the number of opinion labels ($o$), which in our case is 2 or 3 ($o \in \{2,3\}$). Finally, the training is performed by a cross-entropy loss function, and optimized using Adam. For the sake of clarity, Figure \ref{fig_nn} depicts the architecture of the neural network architecture.

\begin{comment}

For the sake of clarity, the mathematical definition of our network is in Equation \ref{eq_nn}, and Figure \ref{fig_nn} depicts the architecture of the neural network.

\begin{equation}
    \begin{split}
        polarity(\textbf{t})   &= \arg \max_{o \in O}(\textbf{y}), \textbf{y} \in {\rm I\!R}^{o}  \\
        \textbf{y} &= \text{softmax}(\textbf{c}), \textbf{c} \in {\rm I\!R}^{o}\\
        \textbf{c} &=\text{ReLu}(\textbf{c}), \textbf{c} \in {\rm I\!R}^{32}\\
        \textbf{c} &= \text{ReLu}(\textbf{c}), \textbf{c} \in {\rm I\!R}^{64}\\
        \textbf{c} &= [\textbf{c}_{\textbf{f}};\textbf{c}_{\textbf{b}}]\\
        \textbf{c}_{\textbf{b}} &= \text{LSTM}^{b}(s_{n:1}), \textbf{s} \in {\rm I\!R}^{l\times 100}\\
        \textbf{c}_{\textbf{f}} &= \text{LSTM}^{f}(s_{1:n}), \textbf{s} \in {\rm I\!R}^{l\times 100}\\
        \textbf{s}   &=  \text{embLookup}(\textbf{t}), \textbf{t} \in {\rm I\!R}^{l}
    \end{split}
    \label{eq_nn}
\end{equation}
%\includegraphics[width=0.8\linewidth, height=10cm]{AuthorKit19/LaTeX/neural_network_transparent.pdf}

\end{comment}

\end{document}